\documentclass[cameraready]{template/mac_automl_xmu_blue}

\makeatletter
\def\input@path{{content/}}
\makeatother
\graphicspath{{content/}}

\usepackage{xargs}
\usepackage{todonotes}
\usepackage{tabularx}
\usepackage{array}
\usepackage{float}
\usepackage{siunitx}
\usepackage{mathtools}
\usepackage{amsthm}

\definecolor{FindingBlue}{RGB}{22,74,126}
\definecolor{FindingBg}{RGB}{247,250,255}
\definecolor{FindingBorder}{RGB}{210,225,242}

\tcbset{
  findingbox/.style={
    enhanced,
    breakable,
    colback=FindingBg,
    colframe=FindingBorder,
    boxrule=0.6pt,
    arc=2.2mm,
    left=2.2mm,
    right=2.2mm,
    top=1.1mm,
    bottom=1.1mm,
    before skip=6pt,
    after skip=6pt,
    fonttitle=\bfseries,
    title={},
    borderline west={1.6pt}{0pt}{FindingBlue},
  },
  appendixbox/.style={
    enhanced,
    breakable,
    colback=white,
    colframe=black!18,
    boxrule=0.4pt,
    arc=1.2mm,
    left=2.0mm,
    right=2.0mm,
    top=1.4mm,
    bottom=1.4mm,
    before skip=6pt,
    after skip=6pt,
    fonttitle=\small\bfseries,
    fontupper=\small,
  },
  appendixcodebox/.style={
    appendixbox,
    colback=black!2,
    fontupper=\ttfamily\small,
    halign=flush left,
  },
}

\newcommand{\best}{\bfseries}
\newcommand{\second}{\itshape}
\newcommand{\na}{\multicolumn{1}{c}{--}}

\usepackage{amsmath,amsfonts,bm}

\def\eqref#1{equation~\ref{#1}}

\def\1{\bm{1}}

\DeclareMathAlphabet{\mathsfit}{\encodingdefault}{\sfdefault}{m}{sl}
\SetMathAlphabet{\mathsfit}{bold}{\encodingdefault}{\sfdefault}{bx}{n}

\setleftheadercontent{%
  \headerlogospace{1.6mm}%
  \headerlogo[-0.08\height]{15.2mm}{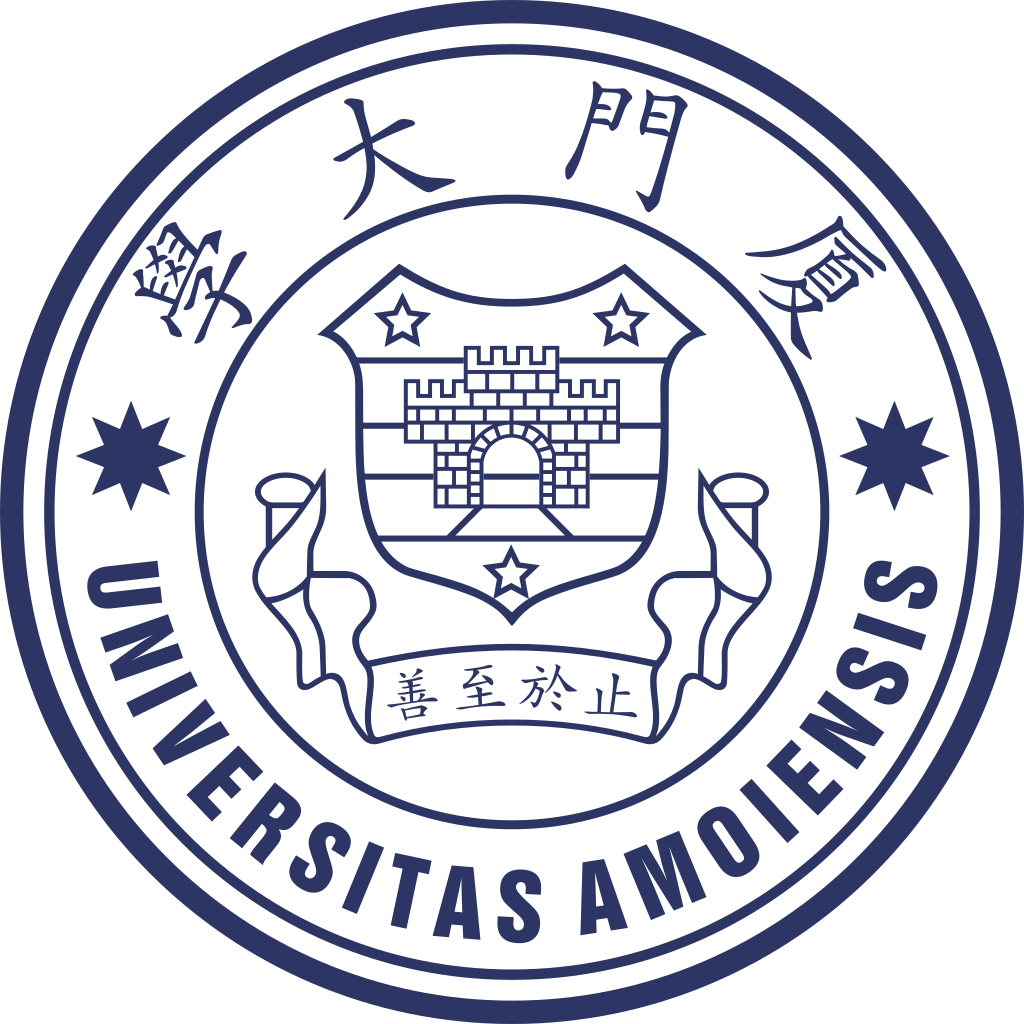}%
}
\setrightheadericon{%
  \headerlogo[-0.05\height]{12.8mm}{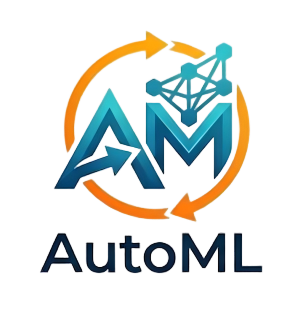}%
  \headerlogospace{1.6mm}%
}
\setrunningheadericon{%
  \headerlogospace{1.2mm}%
  \headerlogo[-0.03\height]{8.2mm}{assets/branding/automl.png}%
}
\setheadergroupname{MAC-AutoML}

\title{Training-Free Multimodal Large Language Model Orchestration}

\setfrontauthors{%
  \authorrow{%
    \authorentry{Tianyu Xie}{1,2},\authorsep
    \authorentry{Yuexiao Ma}{1,2},\authorsep
    \authorentry{Yuhang Wu}{1,2},\authorsep
    \authorentry{Wang Chen}{1,2}}%
  \authorrow{%
    \authorentry{Jiayi Ji}{1,2},\authorsep
    \authorentry{Tat-Seng Chua}{5},\authorsep
    \authorentry{Xiawu Zheng}{1,2,4,\corremailmark},\authorsep
    \authorentry{Rongrong Ji}{1,2,3}}%
}
\setfrontaffiliations{%
  \affiliationline{\authormark{1}Media Analytics and Computing Lab, Xiamen University, Xiamen, China}
  \affiliationline{\authormark{2}Institute of Artificial Intelligence, Xiamen University, Xiamen, China}
  \affiliationline{\authormark{3}School of Informatics, Xiamen University, Xiamen, China}
  \affiliationline{\authormark{4}Peng Cheng Laboratory, Shenzhen, China}
  \affiliationline{\authormark{5}Department of Computer Science, National University of Singapore, Singapore}
}
\setfrontcontact{%
  \contactline{\textbf{\corremailmark\ Corresponding Author: Xiawu Zheng}}
}
\usecustomauthorlayout

\checkdata[Email]{\href{mailto:zhengxiawu@xmu.edu.cn}{{\ifXeTeX\omnimonofont\else\ttfamily\fi zhengxiawu@xmu.edu.cn}}}
\checkdata[Keywords]{Multimodal Large Language Models; Orchestration; Training-Free; Cross-Modal Memory; Unified Interaction}

\abstract{\begin{abstract}
Building interactive omni-modal assistants often relies on end-to-end multimodal alignment to fuse heterogeneous modalities, which incurs substantial data and compute costs and limits extensibility. We present Training-Free Large Language Model Orchestration (LLM Orchestration), a training-free \emph{orchestration} framework that integrates off-the-shelf modality experts into a unified multimodal input--output system without additional gradient-based training for integration. LLM Orchestration comprises three components: (1) an LLM controller that infers user intent and emits explicit control tokens for expert selection and sequencing, enabling protocol-constrained and auditable routing; (2) a text-centric cross-modal memory that compresses multimodal evidence into structured records for lightweight retrieval and reuse, reducing redundant expert invocations across turns; and (3) a unified interaction layer that executes routing and memory decisions to support consistent modality transitions, full-duplex streaming, and interruption-aware dialogue. Across diverse multimodal benchmarks, LLM Orchestration achieves strong performance under standard evaluation constraints while maintaining low orchestration overhead and modular upgradeability, providing a practical alternative to costly joint training for omni-modal systems.
\end{abstract}
}

\begin{document}

\maketitle

\section{Introduction}
Recent advances in Large Language Models (LLMs)~\cite{openai2023gpt4,gemini2023,meta2024llama3,liu2024llavanext} have enabled strong multimodal capabilities. GPT-4o~\cite{hurst2024gpt} further demonstrated the feasibility of processing and generating across multiple modalities within a single assistant, motivating increasing interest in omni-modal systems. Conceptually, omni-modal assistants integrate visual, auditory, and textual evidence to support natural interaction and comprehensive understanding~\cite{zhang2023internlm,chen2024internvl,wang2024qwen2,tong2024cambrian,fu2024vita}. In this paper, we focus on a real-time omni-modal assistant that supports image/video, audio/speech, and text understanding and generation within multi-turn dialogues. A key open question is: how can we build a real-time omni-modal assistant that is \emph{responsive, controllable, and maintainable} as modalities and components evolve?

Despite this progress, many recent multimodal assistants---from vision-language models to emerging omni-modal systems~\cite{zhu2023minigpt,liu2023llava,liu2024llavanext,shin2024x,alayrac2022flamingo,fang2024llama,gemini2023,meta2024llama3,zhao2024humanomni}---largely follow two integration paradigms. \textbf{(A) Unified end-to-end alignment:} a single model is jointly trained or instruction-tuned (often with preference alignment) to handle multiple modalities in one parameter space~\cite{gemini2023,zhao2024humanomni}. \textbf{(B) Backbone expansion with modality modules:} a base LLM is augmented with modality-specific encoders/adapters/projection layers (e.g., image/video/audio modules) and aligned via paired data~\cite{zhu2023minigpt,liu2023llava,alayrac2022flamingo,fang2024llama}. While effective, both paradigms face two practical limitations. \emph{Cost:} aligning heterogeneous modalities typically requires curated datasets and intensive fine-tuning, incurring substantial human effort and compute. \emph{Rigidity:} component upgrades or modality additions often trigger retraining (e.g., swapping an ASR front-end, upgrading a VLM, or adding a new video expert), slowing iteration and complicating deployment. Meanwhile, purely ad-hoc tool-use pipelines often lack system-level guarantees on verification, traceability, and interruption consistency, making them hard to measure and maintain in real-time settings.

Beyond these two paradigms, real-time omni-modal assistants raise a complementary challenge that is not primarily about \emph{representation learning},
but about \emph{execution and interaction}.
In multi-turn, streaming dialogues, the system must make online decisions under strict latency constraints: which computations to trigger,
how to manage intermediate states across modalities, and how to behave under user barge-in and partial outputs.
These requirements are orthogonal to whether the underlying capability is obtained via unified alignment or modular encoders,
yet they critically affect responsiveness, controllability, and maintainability in deployment.
This suggests that, in addition to model integration, we need an explicit orchestration abstraction with well-defined runtime behaviors and measurable guarantees.

To address these limitations, we propose a \textbf{training-free orchestration} framework that composes specialized experts into a single real-time omni-modal assistant. Our key idea is simple: an off-the-shelf LLM controller emits \emph{schema-constrained system tokens} that (i) route each turn to appropriate modality experts and (ii) support interaction control (e.g., interruption), while a text-centric memory \emph{selectively reuses} previously generated multimodal evidence under protocol-enforced rules. 
Concretely, when the executor verifies that the current turn matches a committed evidence key (e.g., identical image hash / segment id) 
and reuse is permitted by runtime policies, it retrieves the corresponding cached record instead of re-invoking the expensive modality expert, 
thereby reducing \emph{unnecessary} expert calls without relying on ambiguous semantic similarity. \textbf{Training-free definition:} throughout this work, ``training-free'' means we perform \emph{no gradient-based training or fine-tuning} for orchestration components (controller, routing logic, memory, and interaction manager); we only use static prompt specifications (token-to-expert mapping and token schema) to integrate off-the-shelf experts. We do not claim training-free multimodal capability; rather, we claim training-free \emph{integration and control} without additional alignment, instruction tuning, or preference optimization for integration.

\paragraph{Positioning.}
Our system composes external experts but differs from generic tool-use agents by defining a \emph{system-level protocol}: closed-vocabulary control tokens coupled with a static token-to-expert mapping (interfaces and I/O contracts). Execution is \emph{router-enforced}: tokens are validated and resolved under explicit runtime policies (timeouts/cancellation), and violations are rejected with fallback routing. The protocol yields a replayable routing trace (token $\rightarrow$ route decision $\rightarrow$ cache-or-call $\rightarrow$ output) and interruption-consistent interaction: on barge-in, in-flight jobs are canceled, ephemeral buffers are separated from committed memory, and only finalized segments/results are persisted.

Our framework comprises three tightly coupled components: \textbf{(C1) Controller-based orchestration}—a central controller emits explicit, auditable tokens to assign each turn to experts; \textbf{(C2) Cross-modal memory integration}—a text-centric memory compresses and retrieves multimodal evidence to preserve context and reduce redundant expert calls; and \textbf{(C3) Unified interaction layer}—a system-level manager executes routing and memory decisions to provide streaming omni-modal I/O with interruption handling. Together, these components offer an interpretable, modular, and extensible alternative to end-to-end joint training.

The main contributions are:
\begin{itemize}
    \item \textbf{(C1) Structured control-token routing.} We introduce an LLM controller that selects and sequences specialized experts via explicit system tokens (e.g., \emph{[S.need\_vision]}, \emph{[S.need\_audio]}, \emph{[S.stop]}), grounded in a fixed token-to-expert mapping and I/O contracts. We report routing latency/overhead and analyze its sensitivity across different expert latency regimes (Figure~\ref{fig:efficiency_analysis}). We further demonstrate extensibility via expert swapping under a fixed controller (Table~\ref{tab:expert_swap_videomme}).

    \item \textbf{(C2) Lightweight cross-modal memory compression and reuse.} We propose a text-centric structured memory that stores expert outputs as compact records and retrieves them using lexical matching with recency signals for bounded worst-case latency and explainable hits. Redundant expert calls are defined by exact evidence-key consistency (e.g., image hash, video chunk id, audio segment timestamp), which avoids ambiguous semantic matching and reduces false reuse. We quantify redundant-call reduction and the resulting savings in Figure~\ref{fig:efficiency_analysis}.

    \item \textbf{(C3) Unified interaction layer for real-time omni-modal I/O.} We design an interaction manager that supports full-duplex streaming and interruption-aware control, where \emph{[S.stop]} cancels in-flight expert/TTS execution and triggers re-routing under updated intent. We evaluate interruption handling with a reproducible test protocol and report cancellation behavior, wasted computation, and end-to-end latency impacts in Section~\ref{sec:exp}.
\end{itemize}

\noindent Under standard benchmark constraints and prompts, our system achieves 44.07\% on WorldSense~\cite{hong2025worldsense} and 69.37\% on MMStar~\cite{chen2024we}. Compared with end-to-end multimodal baselines and our ablated variants (e.g., without memory reuse or protocol-constrained routing), we maintain strong accuracy while keeping routing overhead below 12\% of end-to-end latency and reducing redundant expert invocations in multi-turn interactions (Section~\ref{sec:exp}). We follow official evaluation protocols and scoring scripts, and use experts strictly for modality perception/format conversion without external knowledge access. Detailed accuracy, ablations, and latency breakdowns are provided in Section~\ref{sec:exp}.

\section{Related Work}

\subsection{Training-Based Omni-Modal Systems}
Multimodal Large Language Models  have advanced rapidly via end-to-end multimodal alignment, using either full-parameter training or parameter-efficient adaptation. VITA~\cite{fu2024vita} adopts multi-stage instruction tuning and modality alignment to improve cross-modal coherence. To reduce training cost, parameter-efficient approaches freeze the LLM backbone and train lightweight modality components, e.g., Freeze-Omni~\cite{wang2024freeze}, Mini-Omni2~\cite{xie2024mini}, LLaMA-Omni~\cite{fang2024llama}, and Moshi~\cite{defossez2024moshi}. Recent real-time spoken assistants further incorporate streaming speech understanding and generation, e.g., LLaMA-Omni2~\cite{fang2025llama}.

A recurring limitation is that integration is coupled to specific modality interfaces (e.g., encoders/adapters/projection layers) and paired multimodal data, so component upgrades (e.g., swapping ASR or vision modules) can require additional alignment to preserve calibration and coherence. In contrast, we study \emph{training-free integration and control}: rather than proposing a new omni model, we provide an \emph{orchestration/integration layer} over off-the-shelf experts.

\subsection{Training-Free Routing and Model Composition}
Training-free orchestration composes specialized models via an LLM planner/controller and a tool/model interface. HuggingGPT~\cite{shen2023hugginggpt} uses natural-language tool descriptions with multi-step planning. Visual Programming~\cite{gupta2023visual} and ViperGPT~\cite{suris2023vipergpt} compose visual tools via program synthesis, while ReAct~\cite{yao2023react} interleaves reasoning and action for tool invocation. Chen et al.~\cite{chen2023visualcoord} show that careful prompting can enable training-free coordination of visual modules. Recent work also studies multimodal tool agents that learn to invoke tools under multimodal observations, e.g., MLLM-Tool~\cite{wang2025mllm}.

A related line improves the reliability of tool invocation by enforcing structured outputs, either via constrained generation engines (e.g., grammar-/schema-based decoding)~\cite{dong2025xgrammar} or by learning schema-following policies with supervision or reinforcement learning~\cite{lu2025learning}. These works motivate \emph{parseable, verifiable} action representations, consistent with our closed-vocabulary tokens and router-side validation.

Many prior orchestration systems operate in action spaces that are often open-ended (natural language, code, or free-form arguments), which complicates validation and auditing in low-latency interactive settings. We instead introduce a \emph{protocol-constrained} control-token interface grounded in a static token-to-expert mapping with explicit I/O contracts (Section~\ref{sec:controller_design}); tokens are validated and executed under explicit runtime policies (e.g., timeouts/cancellation). This yields a replayable routing trace (token $\rightarrow$ route decision $\rightarrow$ cache-or-call $\rightarrow$ output), supports measurable routing overhead (Figure~\ref{fig:efficiency_analysis}), and enables explicit rejection of invalid tokens or contract violations.

\subsection{Cross-Modal Memory for Efficient Multi-Turn Orchestration}
Memory mechanisms are widely used in interactive assistants to preserve context and cache intermediate results across turns. Recent work studies how memory structure and retrieval affect agent behaviors and robustness~\cite{zeng2024structural,xu2025mem}. In multimodal orchestration, reuse decisions additionally need to be \emph{verifiable} across modalities and \emph{latency-bounded} for real-time interaction.

We adopt a text-centric, structured cross-modal memory that records expert outputs as compact, reusable records and retrieves them with lightweight rules and recency signals (Section~\ref{sec:memory_integration}). Unlike long-horizon persona/preference memory, we focus on caching modality evidence for multi-turn perception and reasoning. We prioritize \emph{evidence-key consistency} (e.g., image hash, video chunk id, audio segment timestamp) to determine reuse eligibility, favoring verifiability over semantic similarity and mitigating erroneous reuse. We operationalize redundant expert invocations and quantify call reduction via ablations (Figure~\ref{fig:efficiency_analysis}).

\subsection{Interactive Omni-Modal Assistants}
Interactive omni-modal assistants emphasize streaming responsiveness and robustness to user interruptions beyond offline accuracy. VITA~\cite{fu2024vita} studies interruption handling in an end-to-end training setting, while HumanOmni~\cite{zhao2024humanomni} targets human-centric interaction scenarios. Recent spoken assistants further investigate real-time speech understanding/generation with barge-in and interruption handling, e.g., SALMONN-omni~\cite{yu2024salmonn} and LLaMA-Omni2~\cite{fang2025llama}.

Our work is complementary: rather than learning interaction behaviors end-to-end, we provide a unified interaction layer that \emph{operationalizes} routing and memory decisions into explicit execution semantics for real-time pipelines, including canceling in-flight expert/TTS execution and separating ephemeral buffers from committed memory under clear commit rules (Section~\ref{sec:interaction_layer}).

\subsection{Multi-Agent Collaboration Frameworks}
Multi-agent frameworks distribute tasks across multiple interacting LLMs and specialized roles. AutoGen~\cite{wu2023autogen} proposes flexible dialogue patterns for task decomposition. mmctagent~\cite{kumar2024mmctagent} enhances multimodal decision-making via multi-agent coordination, while CrewAI~\cite{duan2024exploration} and TaskWeaver~\cite{qiao2023taskweaver} focus on role-based orchestration and workflow automation. Domain-specific systems such as LawLuo~\cite{sun2024lawluo} simulate professional consultations, and Self-Organized Agents~\cite{ishibashi2024self} and CMAT~\cite{liang2024cmat} explore coordination for code generation and small-model enhancement.

While multi-agent designs can improve flexibility for complex planning, they may incur extra communication overhead and latency variance due to negotiation, which is undesirable in low-latency interactive settings. Our architecture instead adopts a single-controller protocol augmented with cross-modal memory. This reduces coordination overhead and latency variance while retaining explicit and auditable routing decisions (Sections~\ref{sec:controller_design}--\ref{sec:memory_integration}).

These observations motivate an orchestration abstraction that makes \emph{routing}, \emph{reuse}, and \emph{interruption handling} explicit and measurable while remaining modular to component evolution.
Our work takes a step toward this goal by making orchestration and interaction control explicit and measurable in a training-free integration layer over off-the-shelf experts.

\begin{figure*}[t] 
\includegraphics[width=1\textwidth]{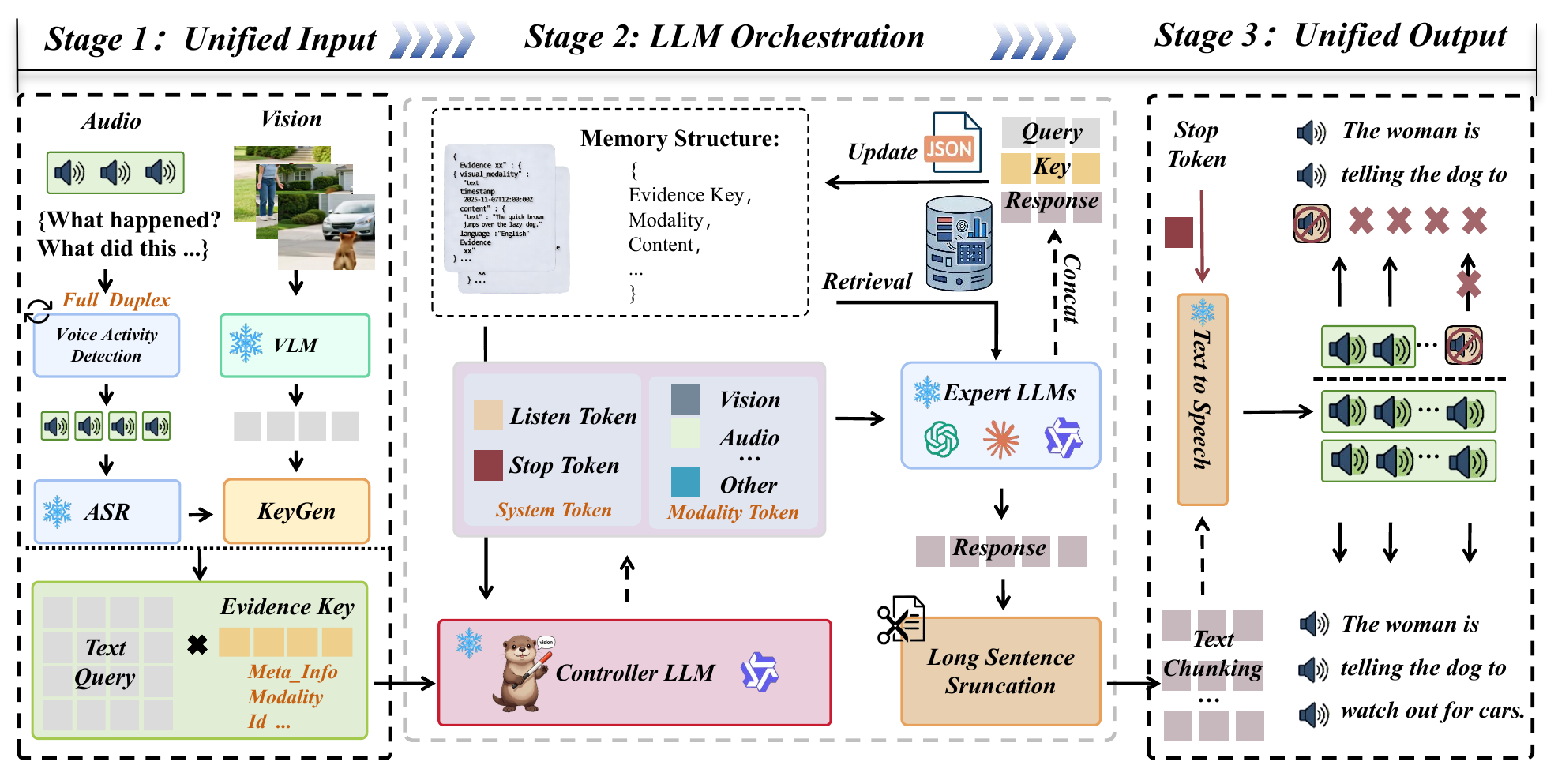} 
\caption{Overview of our training-free MLLM Orchestration pipeline. Stage 1 normalizes multimodal inputs into a unified representation with evidence keys. Stage 2 uses a controller to emit closed-vocabulary control tokens, routes to external experts, and performs cache-or-call retrieval with evidence-keyed memory. Stage 3 streams ordered outputs and supports barge-in by canceling pending jobs.} 
\label{fig:01_compare} 
\end{figure*}

\section{Method}
\label{sec:method}

\subsection{Motivation and Design Goal}
Recent omni-modal assistants achieve strong interactive capabilities by aligning audio, vision, and language through substantial training (e.g., VITA~\cite{fu2024vita}).%
\footnote{A common abstraction for such systems is gradient-based multimodal alignment on paired observations, e.g.,
$\theta^{\star}=\arg\min_{\theta}\mathbb{E}_{(a,v,y)\sim\mathcal{D}}\big[\mathcal{L}_{\text{align}}(f_{\theta}^{a}(a), f_{\theta}^{v}(v), f_{\theta}^{y}(y))\big]$,
where $a$ and $v$ denote audio and visual inputs, $y$ denotes language supervision, and $\mathcal{L}_{\text{align}}$ enforces cross-modal consistency under an end-to-end or staged recipe.}
While effective, alignment-centric designs are costly and can tightly couple components, making upgrades (e.g., swapping ASR/LLM/TTS) difficult without re-tuning.

We ask whether omni-style real-time interaction can be achieved with off-the-shelf experts and explicit runtime control, rather than requiring a newly trained omni model. Our key observation is that real-time omni behavior hinges on \emph{interaction-time control}---when to listen vs.\ respond, when to stop on user barge-in, and which modality evidence to acquire---beyond representation learning alone.

We therefore target \textbf{training-free integration and control}: the orchestration layer (routing, memory reuse, streaming, and interruption semantics) is specified by a fixed protocol and deterministic execution rules, with no gradient-based training for integration components. Concretely, an LLM controller emits closed-vocabulary state/modality tokens that a deterministic router executes to invoke or reuse pluggable experts, while an evidence-keyed memory enables verifiable cache-or-call reuse across turns. Our objective is to match omni interaction behavior through modular orchestration rather than learned multimodal alignment.

\subsection{Stage 1: Unified Input}
Stage~1 runs continuously and normalizes heterogeneous inputs into two outputs for the next stage: a consolidated query $q_t$ and a set of evidence references $\mathcal{K}_t$.
For speech, Voice Activity Detection (VAD) segments the microphone stream into intervals and ASR transcribes each interval into text tokens.
We denote the resulting transcript evidence as $text_{\text{audio}}$; Stage~1 buffers/merges these tokens and consolidates them into $q_t$ (with timestamps retained for bookkeeping).

For non-text inputs, Stage~1 does not expose raw payloads to the controller; instead it assigns each segment an \emph{evidence key} using system code (\textsc{KeyGen}) and returns the keys as references.
An evidence key is a system-generated identifier that uniquely names a concrete multimodal segment; Stage~2 refers to raw inputs only through these keys and never directly accesses the payload:
\begin{equation}
k=\textsc{KeyGen}(m,s)\triangleq \langle m,\ s,\ h\rangle,
\label{eq:keygen}
\end{equation}
where $m$ is the modality tag, $s$ is a segment identifier (audio timestamp span, image fingerprint, or video chunk/frame id), and $h$ is an integrity checksum.
Stage~1 maintains a local mapping from $k$ to the raw payload pointer; the controller only observes $\mathcal{K}_t$.

For images/videos, Stage~1 always registers segments and outputs their keys, but vision-to-text evidence is computed only on demand:
the vision expert is silent by default and is invoked only after Stage~2 requests vision (e.g., \texttt{vision}); upon activation, Stage~1 produces a textualized description $text_{\text{vision}}$ for the requested key(s).
During system playback, any new valid VAD segment is flagged as a barge-in event and reported to Stage~2 for decision making.

\subsection{Stage 2: LLM Orchestration}
\label{sec:controller_design}
Stage~2 is the decision and execution core.
At turn $t$, it takes $(q_t,\mathcal{K}_t)$ from Stage~1, together with interaction events (e.g., barge-in), and a bounded memory view $\hat{M}_{t-1}$ derived from its own session memory $M_{t-1}$.
Stage~2 then produces either a control decision (keep listening / stop output) or a response text $o_t^{\text{text}}$ prepared for streaming.

\textbf{Controller $\rightarrow$ Router $\rightarrow$ Experts.}
We use an off-the-shelf LLM as a controller.
Given a fixed system prompt that specifies the closed token vocabulary and the current interaction state, the controller emits a short sequence of control tokens indicating (i) system status (listen/stop/respond) and (ii) which modality evidence is required (e.g., \texttt{vision} for some referenced key).
A deterministic router validates these tokens and resolves them through a system-side mapping $\mathcal{E}$, which binds each routing token to an expert endpoint, required inputs, output schema, and runtime policies (timeouts/cancellability).
This step selects the concrete expert(s) to invoke and constructs their inputs by combining (a) the user-side query $q_t$, (b) the referenced evidence keys in $\mathcal{K}_t$ (used to locate raw payloads outside the controller), and (c) controller-organized textual context from $\hat{M}_{t-1}$ when needed.
Each invoked expert then returns textualized evidence (as text tokens) that is schema-validated by the router before being used downstream.

\textbf{Evidence-keyed memory and cache-or-call.}
\label{sec:memory_integration}
To reduce redundant expert invocations across turns, Stage~2 maintains an evidence-keyed memory $M$ as a per-session store of \emph{textualized evidence records} indexed by evidence key $k$.
A record contains a compact, schema-valid summary of an expert output (e.g., ASR transcripts $text_{\text{audio}}$, vision descriptions $text_{\text{vision}}$), together with lightweight metadata (modality tag, timestamps).
Memory stores \emph{evidence}, not raw payloads: raw audio/images/videos remain outside the controller and are accessible only to experts via evidence references.
To bound controller context length, Stage~2 exposes only a compressed view $\hat{M}_{t-1}=\textsc{Compress}(M_{t-1})$ in the prompt.

When evidence is required (e.g., \texttt{vision} for $k\in\mathcal{K}_t$), Stage~2 applies the deterministic rule:
\begin{equation}
d(k)=
\begin{cases}
\textsc{Read}(M_{t-1},k), & \text{if }k\in M_{t-1},\\
\textsc{CallExpert}(k)\ \text{and write to }M_t, & \text{otherwise.}
\end{cases}
\label{eq:cachecall}
\end{equation}
Key equality is the default reuse criterion, making reuse decisions verifiable.
In particular, vision textualization is gated by control tokens: when \texttt{vision} is present, the selected vision expert is invoked on the referenced key(s) and produces $text_{\text{vision}}$; if the corresponding record already exists in memory, the expert call is skipped.
Analogously, when \texttt{need\_audio} is present, the ASR/audio expert produces (or reuses) $text_{\text{audio}}$.

\textbf{Response synthesis and streaming preparation.}
After evidence acquisition, Stage~2 concatenates (i) the consolidated query $q_t$, (ii) retrieved memory records relevant to $\mathcal{K}_t$ (via $\hat{M}_{t-1}$), and (iii) any newly produced expert evidence into a single textual context (Concat in Figure~\ref{fig:01_compare}), applies a lightweight truncation/compression rule to bound context length, and generates the response text $o_t^{\text{text}}$ with a text-only generator.
To prepare for real-time output, Stage~2 further segments $o_t^{\text{text}}$ into speakable chunks using punctuation and a maximum-token constraint; these chunks (with monotonic chunk IDs) are then passed to Stage~3 for parallel Text-to-Speech (TTS) and ordered playback.
Finally, Stage~2 commits the response and the evidence keys used as a structured session record in $M_t$ and logs a replayable trace (tokens $\rightarrow$ routes $\rightarrow$ cache-or-call $\rightarrow$ outputs).

\subsection{Stage 3: Streaming Output with Barge-in Cancellation}
\label{sec:interaction_layer}
Stage~3 takes the response text $o_t^{\text{text}}$ and control decisions (e.g., \texttt{stop}) from Stage~2 and realizes them as ordered streaming I/O.
When a response is to be spoken, Stage~3 chunks $o_t^{\text{text}}$ by punctuation and a maximum length, assigns monotonic chunk IDs, and synthesizes chunks asynchronously with a fixed-size worker pool.
Chunks may finish out of order, but playback is strictly ordered by chunk ID.

Interruption is enforced by cancellation.
When Stage~2 emits \texttt{stop} (e.g., after receiving a barge-in event), Stage~3 cancels pending/running TTS jobs and clears both the generation queue and playback buffer, guaranteeing silence after interruption.
We apply the same cancellation policy to expert calls marked cancellable in $\mathcal{E}$, ensuring consistent interruption behavior across perception and rendering.

\subsection{Multi-turn Full-Duplex Workflow}
Algorithm~\ref{alg:multiturn-workflow} summarizes the multi-turn interaction loop, including barge-in interruption, deterministic planning, evidence acquisition via cache-or-call, streaming TTS, and commit-on-complete memory updates.
We provide full workflow pseudocode in Appendix~\ref{sec:algorithm} for readability.

\subsection{Pipeline Overview}
\paragraph{Running example (Figure~\ref{fig:01_compare}).}
Consider a user who speaks ``What happened?'' while providing a short video clip where a dog runs as a car approaches.
In Stage~1, VAD+ASR transcribes the utterance into $text_{\text{audio}}$ and consolidates it into the query $q_t$; the video is registered with an evidence key $k_{\text{vid}}\in\mathcal{K}_t$ via \textsc{KeyGen} (without running vision by default).
In Stage~2, the controller, conditioned on the system prompt and $(q_t,\mathcal{K}_t,\hat{M}_{t-1})$, emits \texttt{respond} with \texttt{vision} for $k_{\text{vid}}$; the router resolves this token to the vision expert, applies cache-or-call on $k_{\text{vid}}$, and obtains (or reuses) a textualized description $text_{\text{vision}}$ (e.g., the dog notices the car and moves away).
The generator then produces the response text (e.g., ``The woman is telling the dog to watch out for cars.'') and Stage~2 segments it into speakable chunks for Stage~3.
Finally, Stage~3 performs parallel Text-to-Speech (TTS) on the chunks and plays them in order; if the user barges in mid-playback, Stage~2 emits \texttt{stop} and Stage~3 cancels pending jobs and clears buffers.

\section{Experiments and Results}
\label{sec:exp}

We evaluate our training-free orchestration along three system-level claims: (C1) protocol-constrained routing for robust control and modular composition, (C2) evidence-keyed memory for reuse and efficiency, and (C3) an interaction layer for streaming and interruption-consistent runtime behavior. We report end-to-end benchmark results in Table~\ref{tab:omni_comparison} and isolate each claim via controller/expert swapping, memory ablations, and compute-only latency breakdowns (Figure~\ref{fig:efficiency_analysis}, Table~\ref{tab:expert_swap_videomme}).

\subsection{Experimental Setup}

\noindent\textbf{Controller \& Expert Configuration.}
We employ Qwen2.5-14B~\cite{chu2024qwen2} as the central controller LLM with deterministic decoding (temperature=0.0, top-p=1.0) to ensure stable control-token emission. The system exposes a set of \emph{pluggable} experts accessed via a system-side token-to-expert mapping. The controller emits closed-vocabulary state/modality tokens (e.g., \texttt{need\_vision}, \texttt{listen}, \texttt{stop}); experts are invoked only when routed by these tokens. When no expert is required, the controller performs text-only reasoning. In our Video-MME pipeline, we primarily use frame-based visual experts (e.g., Qwen2.5-VL) under a fixed sampling budget, while dedicated video experts (e.g., LLaVA-Video) can be plugged in when temporal modeling is desired. Unless otherwise stated, the reported Video-MME results use frame-based experts plus ASR transcripts (Section~\ref{sec:exp_video}). All local experts use greedy decoding (temperature=0.0) unless otherwise noted. We run inference on 8$\times$NVIDIA A100 GPUs (80GB) with mixed precision (FP16).

\noindent\textbf{Baselines.}
We compare against: (i) \textit{single-model open-weight MLLMs} (e.g., Qwen2.5-VL, InternVL-2, LLaVA-OV), (ii) \textit{jointly trained omni-models} (e.g., Qwen2.5-Omni~\cite{Qwen2.5-Omni}, VITA~\cite{fu2024vita}, M2-omni~\cite{guo2025m2}), and (iii) \textit{commercial APIs} (GPT-4o~\cite{hurst2024gpt}, Claude 3.5~\cite{claude2024}, Gemini-1.5-Pro~\cite{team2024gemini}) as reference points. For our system, Table~\ref{tab:omni_comparison} reports the average active visual expert parameters under dynamic routing; the controller is fixed at 14B.
\textit{Training-free} means we perform no additional joint end-to-end training or cross-modal alignment to obtain the unified assistant; experts are off-the-shelf pretrained/aligned models coordinated via prompts and explicit control tokens.

\noindent\textbf{Evaluation Protocol.}
We evaluate on established benchmarks covering: general multimodal understanding (MME~\cite{fu2023mme}, MMBench-EN/CN~\cite{liu2024mmbench}), challenging vision QA (MMStar~\cite{chen2024we}, MMMU~\cite{yue2024mmmu}), long video understanding (LVBench~\cite{wang2024lvbench}, Video-MME~\cite{fu2024video}), holistic omni-modal evaluation (WorldSense~\cite{hong2025worldsense}), and specialized domains (MathVision~\cite{wang2024measuring}, CC-OCR~\cite{yang2024cc}). We follow official evaluation scripts when provided and fix the random seed (seed=42). For video benchmarks, we standardize frame sampling to FPS=20 and cap processing to \emph{MAX\_FRAMES\_PER\_BATCH}=20 unless a benchmark protocol constrains otherwise.

\noindent\textbf{Latency Measurement.}
We report \emph{compute-only} latency for local models, including controller inference, token parsing/validation, memory lookup, scheduling decisions, and expert inference, excluding network I/O. Timing statistics are computed over 100 randomly sampled queries per benchmark; we additionally report system transmission overhead (e.g., IPC) separately when relevant.

\subsection{Comparison with Mainstream Omni Models}
\label{sec:exp_omni}

This subsection reports end-to-end comparisons against mainstream multimodal/omni models. Table~\ref{tab:omni_comparison} summarizes unified results across general multimodal, vision, and video benchmarks.

\noindent\textbf{Main Results (Table~\ref{tab:omni_comparison}).}
Table~\ref{tab:omni_comparison} summarizes end-to-end performance across representative multimodal, vision, long-video, and holistic omni settings. Overall, our orchestration is competitive with strong open-weight baselines and approaches natively trained omni models in aggregate. We attribute the gains to protocol-driven expert routing and explicit evidence composition (e.g., combining visual evidence with ASR transcripts), without requiring joint end-to-end training for integration.

\begin{figure}[!t]
  \centering
  \includegraphics[width=\linewidth]{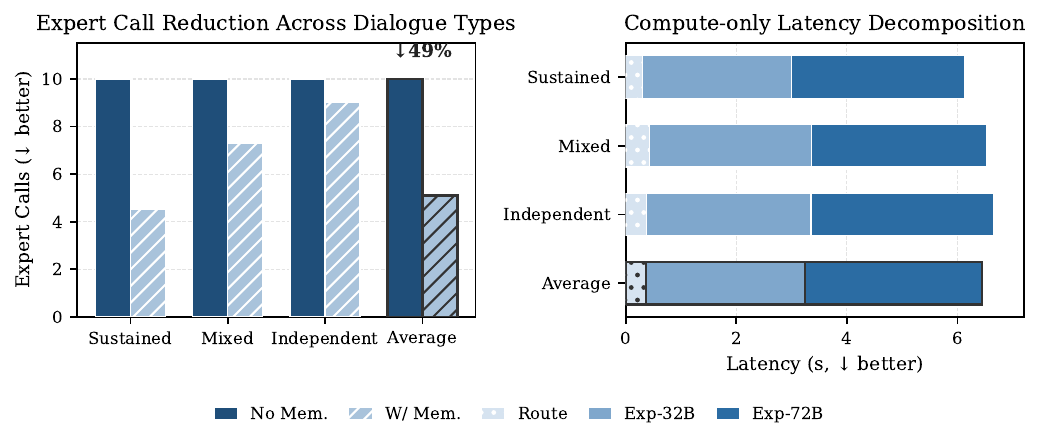}
  \caption{Efficiency analysis. Left: expert-call reduction across dialogue types when enabling cross-modal memory (avg.\ $49\%$). Right: compute-only latency decomposition; expert inference dominates while routing and orchestration overhead remain small. Here, $T_{\text{route}}$ includes controller inference, token parsing/validation, memory lookup (when enabled), and scheduling, and the same measurement scope is used across settings for a fair overhead comparison. Overhead is computed as $100\times \frac{T_{\text{route}}}{T_{\text{route}}+T_{\text{Exp-72B}}}$ and stays below $12\%$ across scenarios (avg.\ $10.36\%$).}
  \label{fig:efficiency_analysis}
\end{figure}

\noindent\textbf{Efficiency (Avg. Inference Latency).}
For methods with reported timing, latency reflects the usual accuracy--speed trade-off. Our system operates in the same order of magnitude as fast open-weight baselines, while preserving competitive accuracy (Table~\ref{tab:omni_comparison}). Detailed latency decomposition and overhead are reported in Figure~\ref{fig:efficiency_analysis}.

\begin{table*}[!t]
	  \centering
	  \renewcommand{\arraystretch}{1.06}
	  \footnotesize
	  \setlength{\tabcolsep}{0.9mm}
	  \resizebox{\textwidth}{!}{%
		  \begin{tabular}{@{}l c S[table-format=4.1] S[table-format=2.2] S[table-format=2.2] S[table-format=2.2] S[table-format=2.2] S[table-format=2.2] S[table-format=2.2] S[table-format=1.1]@{}}
		    \toprule
		    \textbf{Model} & \textbf{Avg. Active Params} & \multicolumn{3}{c}{\textbf{General Multimodal}} & \multicolumn{4}{c}{\textbf{Vision Understanding}} & \textbf{Efficiency} \\
		    \cmidrule(lr){3-5}\cmidrule(lr){6-9}\cmidrule(lr){10-10}
		    & \textbf{(B)} & \multicolumn{1}{c}{MME} & \multicolumn{1}{c}{MMBench-EN} & \multicolumn{1}{c}{MMStar}
		    & \multicolumn{1}{c}{LVBench} & \multicolumn{1}{c}{MMMU} & \multicolumn{1}{c}{Video-MME} & \multicolumn{1}{c}{WorldSense} & \multicolumn{1}{c}{Time (s)} \\
		    \midrule
			    Qwen2.5-VL   & 7 & 1673 & 84.45 & 59.94 & 45.30 & \na & 56.62 & \na & \na \\
			    Qwen2.5-VL   & 32 & 1915 & 85.55 & 66.43 & \second 49.00 & \na & 62.39 & \na & \na \\
			    Qwen2.5-VL   & 72 & 1980 & \second 86.61 & \second 68.22 & 47.30 & \na & 65.74 & \na & \na \\
			    Qwen-VL-Max  & \multicolumn{1}{c}{-} & 2281 & 77.60 & \na & \na & \na & 51.30 & \na & \na \\
				Qwen2.5-Omni & 7 & \best 2340 & 81.80 & 64.00 & \na & \second 59.20 & 64.30 & \second 45.4 & 6.0 \\
				VITA1.5      & 7 & 2006 & 71.80 & 46.40 & \na & 47.30 & 59.20 & 36.9 & 3.7 \\
				LLaVA-OV     & 7 & \na & 80.80 & 61.70 & \na & \na & 58.20 & 37.7 & \na \\
				LLaVA-OV     & 72 & \na & 85.90 & 66.10 & 26.90 & \na & 66.20 & \na & \na \\
				InternVL-2   & 8 & \na & 81.70 & 59.40 & \na & \na & \na  & 39.1 & \na \\
				InternVL-2   & 26 & \na & 83.40 & 60.40 & \na & \na & \na & \na & \na \\
				IXC2.5       & 7 & \na & \na & \na & \na & \na & 60.60 & \na & \na \\
				M2-omni      & 9 & \na & \na & 60.50 & \na & 51.20 & 60.40 & \na & \na \\
				Gemini-1.5-Pro & \multicolumn{1}{c}{-} & \na & \na & \na & 33.10 & \na & \best 75.00 & \best 48.0 & \na \\
				GPT-4V       & \multicolumn{1}{c}{-} & 1409 & 75.00 & 57.10 & \na & \na & 59.90 & \na & \na \\
				GPT-4o       & \multicolumn{1}{c}{-} & \second 2310 & 83.10 & \second 64.70 & 34.70 & \second 59.20 & \second 71.90 & 42.6 & 1.2 \\
		\rowcolor{gray!12} \textbf{Ours:base Qwen2.5} & \multicolumn{1}{c}{53.8} & 1922 & \best 88.54 & \best 69.37 & \best 50.27 & \best 70.04 & 65.58 & 44.1 & 3.2 \\
		    \bottomrule
		  \end{tabular}
	  }
	  \caption{Unified comparison on general multimodal and vision/video benchmarks. \textbf{Avg. Active Params (B)} reports dense model size for single-model baselines and the average active visual expert budget for our routed system; our controller is fixed at 14B and not included in this column. ``--'' indicates unavailable or undisclosed information. \textbf{WorldSense} measures holistic omni-modal understanding. Best/second-best marked by \best/\second. Latency and overhead analyses are reported in Figure~\ref{fig:efficiency_analysis}.}
	  \label{tab:omni_comparison}
\end{table*}

\begin{tcolorbox}[findingbox]
\textbf{Finding A (Competitive end-to-end performance).}
Across representative benchmarks, protocol-driven orchestration with off-the-shelf experts achieves competitive end-to-end accuracy, indicating that strong omni-modal capability can be obtained without joint end-to-end training for integration (Table~\ref{tab:omni_comparison}).
\end{tcolorbox}

\noindent\textbf{(C1) Controller Replacement under a Fixed Token Protocol.}
Under the same token protocol, prompts, deterministic decoding, and runtime semantics, we replace the controller while keeping the expert routing and execution unchanged. The resulting accuracy is similar, suggesting routing behavior is primarily constrained by the protocol-defined action space rather than a specific controller implementation.

\noindent\textbf{(C2) Memory Integration Efficiency.}
We ablate memory by disabling retrieval and forcing recomputation at each turn under the same protocol and runtime semantics. Memory substantially reduces redundant expert invocations, with larger gains on sustained-query dialogues where evidence is repeatedly referenced (Figure~\ref{fig:efficiency_analysis}).

\begin{tcolorbox}[findingbox]
\textbf{Finding B (Modularity and efficiency with low overhead).}
Under a fixed token protocol and runtime semantics, controller/expert swapping preserves behavior while accuracy scales with expert capability (Table~\ref{tab:expert_swap_videomme}). Evidence-keyed memory reduces redundant expert calls (Figure~\ref{fig:efficiency_analysis}), and routing/scheduling overhead remains a small fraction of total compute.
\end{tcolorbox}

\noindent\textbf{(C1/C3) Orchestration Overhead Analysis.}
Figure~\ref{fig:efficiency_analysis} shows that expert forward passes dominate compute-only latency, while routing (controller inference, token parsing, memory lookup, scheduling) adds limited overhead. We additionally measure a small transmission overhead (excluded from compute-only time) and report it separately.

\subsection{Video Understanding Performance and Holistic Evaluation}
\label{sec:exp_video}

\begin{figure}[!ht]
    \centering
    \includegraphics[width=0.95\linewidth]{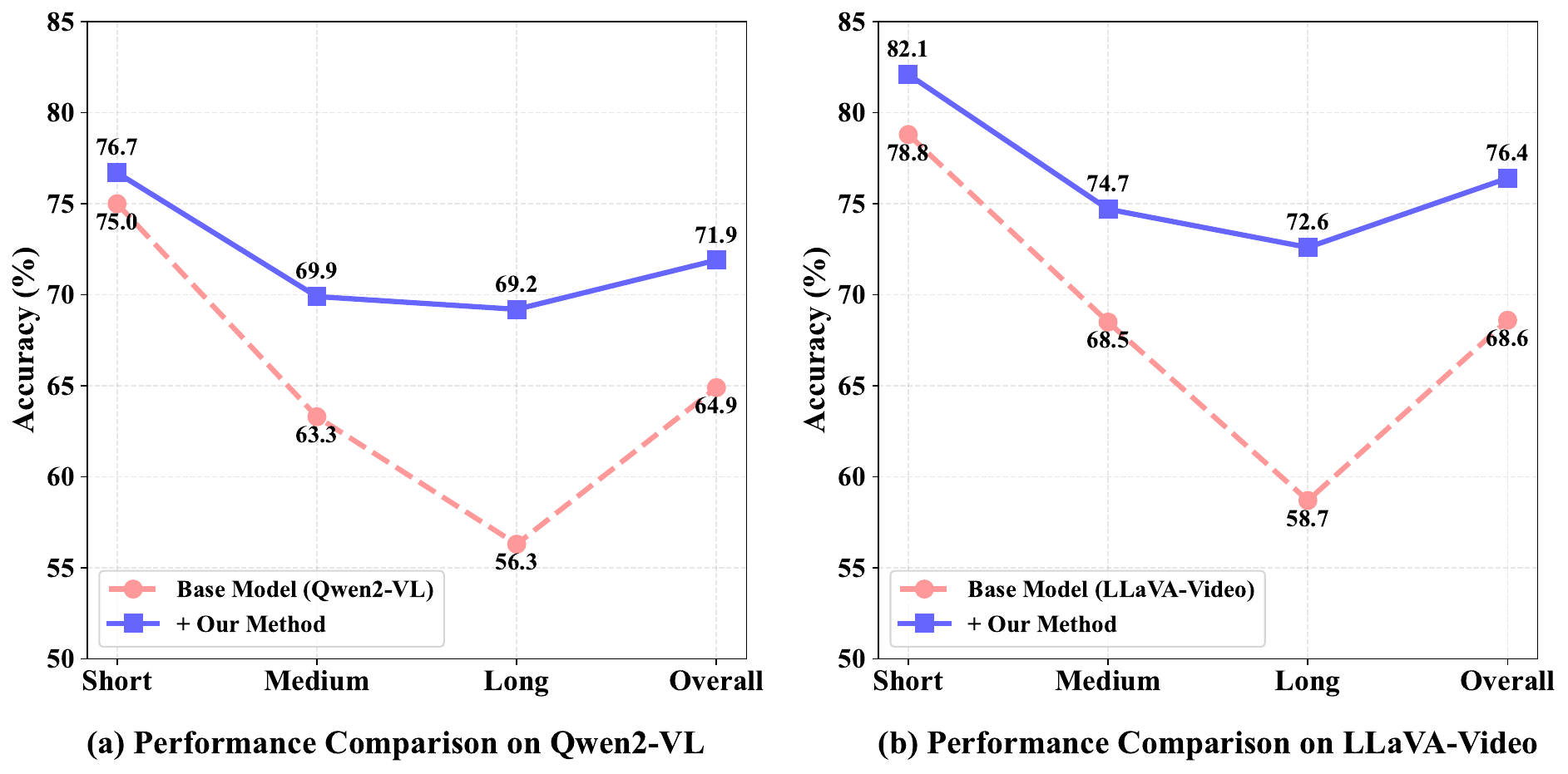}
    \caption{Video-MME accuracy stratified by video length. Our orchestration consistently improves over the corresponding single-model baseline under the same backbone, with larger gains on longer videos, indicating that protocol-driven routing and evidence aggregation are particularly beneficial when temporal context grows. (a) Qwen2-VL backbone. (b) LLaVA-Video backbone.}
    \label{fig:video_performance}
\end{figure}

We evaluate audio-visual understanding under a standardized frame-sampling budget with ASR transcripts as textual audio evidence. Protocol-driven routing enables selective expert invocation and explicit evidence aggregation, which is particularly helpful as temporal context grows under sparse sampling. Stratified results by video length are summarized in Figure~\ref{fig:video_performance}.

\noindent\textbf{Expert Swapping Ablation (Video-MME).}
To verify extensibility, we keep the controller fixed and swap the frame-based visual and reasoning experts under identical evaluation settings. With Qwen2.5-VL-32B + Qwen2.5-VL-72B, we obtain \textbf{65.58\%}. Replacing them with Qwen3-VL-32B-Instruct + Qwen3-VL-235B-A22B-Instruct \cite{yang2025qwen3} improves accuracy to \textbf{72.31\%} (same ASR pipeline and frame sampling: 20 FPS. This ablation demonstrates expert extensibility: keeping the controller and protocol fixed, upgrading the expert pair improves accuracy while preserving the same orchestration semantics (Table~\ref{tab:expert_swap_videomme}).
\begin{table}[!t]
  \centering
  \renewcommand{\arraystretch}{1}
  \scriptsize
  \setlength{\tabcolsep}{1mm}
  \caption{Video-MME under a fixed controller: expert-pair swapping (20 FPS, \emph{MAX\_FRAMES\_PER\_BATCH}=20).}
  \label{tab:expert_swap_videomme}
  \begin{tabular}{@{}>{\raggedright\arraybackslash}m{0.30\columnwidth} >{\centering\arraybackslash}m{0.30\columnwidth} >{\centering\arraybackslash}m{0.30\columnwidth}@{}}
    \toprule
    \textbf{Expert pair} & \textbf{Acc (\%)} & \textbf{Usage (V/R, \%)} \\
    \midrule
    \shortstack[l]{\textbf{V:} Qwen2.5\allowbreak -VL\allowbreak -32B\\[-0.25ex]\textbf{R:} Qwen2.5\allowbreak -VL\allowbreak -72B} & 65.58 & 45.62/54.38 \\
    \shortstack[l]{\textbf{V:} Qwen3\allowbreak -VL\allowbreak -32B\\[-0.25ex]\textbf{R:} Qwen3\allowbreak -VL\allowbreak -235B\allowbreak -A22B} & 72.31 & 43.49/56.51 \\
    \bottomrule
\end{tabular}
\end{table}

We further evaluate holistic omni-modal capability on WorldSense~\cite{hong2025worldsense}. On WorldSense, our orchestration remains competitive under a holistic omni setting (Table~\ref{tab:omni_comparison}). The task-type breakdown (Appendix~\ref{sec:appendix_additional_results}, Table~\ref{tab:worldsense_breakdown}) suggests a temporal-resolution trade-off under sparse sampling: high-level semantic/causal reasoning benefits from modular evidence integration, while fine-grained temporal counting remains challenging.

\noindent\textbf{Limitations and Trade-offs.}
Fine-grained temporal counting remains challenging with sparse frame sampling. This reflects an accuracy--resolution trade-off: modular evidence helps high-level reasoning, whereas dense temporal encoders better capture fast dynamics.

\section{Conclusions}
\label{sec:con}

We presented a training-free LLM orchestration framework that composes off-the-shelf multimodal experts into a unified full-duplex assistant through a protocol-constrained control interface. A central controller emits closed-vocabulary state/modality tokens that are deterministically executed by the runtime router, enabling auditable expert invocation, interruption-consistent streaming (e.g., cancellation and safe fallbacks), and modular extensibility without joint end-to-end training for integration. We further introduced an evidence-keyed, text-centric cross-modal memory that supports multi-turn reuse via cache-or-call and bounded compression, improving efficiency while keeping orchestration overhead small. Experiments across heterogeneous benchmarks and system ablations show that the protocol largely decouples orchestration from expert implementations, so controller/expert upgrades can translate into capability gains without retraining.

\noindent\textbf{Future work.}
Current limitations include sparse temporal sampling for fine-grained events and non-oracle routing in ambiguous cases. Future work will study higher-resolution evidence retention, specialized temporal experts, and lightweight routing refinement while preserving protocol interpretability. Taken together, this work provides a scalable recipe for building real-time omni-modal assistants from evolving experts while keeping execution semantics explicit.

\section*{Acknowledgements}
This work is supported by the National Key Research and Development Program of China (No. 2025YFE0113500),  the National Natural Science Foundation of China ( No. 62576299) and the Fundamental Research Funds for the Central Universities.

\clearpage
\bibliographystyle{plainnat}
\bibliography{references}

\begin{thebibliography}{53}
\providecommand{\natexlab}[1]{#1}
\providecommand{\url}[1]{\texttt{#1}}
\expandafter\ifx\csname urlstyle\endcsname\relax
  \providecommand{\doi}[1]{doi: #1}\else
  \providecommand{\doi}{doi: \begingroup \urlstyle{rm}\Url}\fi

\bibitem[Alayrac et~al.(2022)Alayrac, Donahue, Luc, Miech, Barr, Hasson, Lenc, Mensch, Millican, Reynolds, et~al.]{alayrac2022flamingo}
Jean-Baptiste Alayrac, Jeff Donahue, Pauline Luc, Antoine Miech, Iain Barr, Yana Hasson, Karel Lenc, Arthur Mensch, Katherine Millican, Malcolm Reynolds, et~al.
\newblock Flamingo: a visual language model for few-shot learning.
\newblock \emph{Advances in neural information processing systems}, 35:\penalty0 23716--23736, 2022.

\bibitem[Anthropic(2024)]{claude2024}
Anthropic.
\newblock Claude-3.5.
\newblock \url{https://www.anthropic.com/news/claude-3-5-sonnet}, 2024.
\newblock Accessed: 2024-02-11.

\bibitem[Chen et~al.(2023)Chen, Li, Shen, Yang, Li, Keutzer, Darrell, and Liu]{chen2023visualcoord}
Liangyu Chen, Bo~Li, Sheng Shen, Jingkang Yang, Chunyuan Li, Kurt Keutzer, Trevor Darrell, and Ziwei Liu.
\newblock Large language models are visual reasoning coordinators.
\newblock In \emph{Advances in Neural Information Processing Systems (NeurIPS)}, 2023.

\bibitem[Chen et~al.(2024{\natexlab{a}})Chen, Li, Dong, Zhang, Zang, Chen, Duan, Wang, Qiao, Lin, et~al.]{chen2024we}
Lin Chen, Jinsong Li, Xiaoyi Dong, Pan Zhang, Yuhang Zang, Zehui Chen, Haodong Duan, Jiaqi Wang, Yu~Qiao, Dahua Lin, et~al.
\newblock Are we on the right way for evaluating large vision-language models?
\newblock \emph{arXiv preprint arXiv:2403.20330}, 2024{\natexlab{a}}.

\bibitem[Chen et~al.(2024{\natexlab{b}})Chen, Wu, Wang, Su, Chen, Xing, Zhong, Zhang, Zhu, Lu, et~al.]{chen2024internvl}
Zhe Chen, Jiannan Wu, Wenhai Wang, Weijie Su, Guo Chen, Sen Xing, Muyan Zhong, Qinglong Zhang, Xizhou Zhu, Lewei Lu, et~al.
\newblock Internvl: Scaling up vision foundation models and aligning for generic visual-linguistic tasks.
\newblock In \emph{Proceedings of the IEEE/CVF conference on computer vision and pattern recognition}, pages 24185--24198, 2024{\natexlab{b}}.

\bibitem[Chu et~al.(2024)Chu, Xu, Yang, Wei, Wei, Guo, Leng, Lv, He, Lin, et~al.]{chu2024qwen2}
Yunfei Chu, Jin Xu, Qian Yang, Haojie Wei, Xipin Wei, Zhifang Guo, Yichong Leng, Yuanjun Lv, Jinzheng He, Junyang Lin, et~al.
\newblock Qwen2-audio technical report.
\newblock \emph{arXiv preprint arXiv:2407.10759}, 2024.

\bibitem[D{\'e}fossez et~al.(2024)D{\'e}fossez, Mazar{\'e}, Orsini, Royer, P{\'e}rez, J{\'e}gou, Grave, and Zeghidour]{defossez2024moshi}
Alexandre D{\'e}fossez, Laurent Mazar{\'e}, Manu Orsini, Am{\'e}lie Royer, Patrick P{\'e}rez, Herv{\'e} J{\'e}gou, Edouard Grave, and Neil Zeghidour.
\newblock Moshi: a speech-text foundation model for real-time dialogue.
\newblock \emph{arXiv preprint arXiv:2410.00037}, 2024.

\bibitem[Dong et~al.(2025)Dong, Ruan, Cai, Xu, Zhao, Lai, and Chen]{dong2025xgrammar}
Yixin Dong, Charlie~F Ruan, Yaxing Cai, Ziyi Xu, Yilong Zhao, Ruihang Lai, and Tianqi Chen.
\newblock Xgrammar: Flexible and efficient structured generation engine for large language models.
\newblock \emph{Proceedings of Machine Learning and Systems}, 7, 2025.

\bibitem[Duan and Wang(2024)]{duan2024exploration}
Zhihua Duan and Jialin Wang.
\newblock Exploration of llm multi-agent application implementation based on langgraph+ crewai.
\newblock \emph{arXiv preprint arXiv:2411.18241}, 2024.

\bibitem[Fang et~al.(2024)Fang, Guo, Zhou, Ma, Zhang, and Feng]{fang2024llama}
Qingkai Fang, Shoutao Guo, Yan Zhou, Zhengrui Ma, Shaolei Zhang, and Yang Feng.
\newblock Llama-omni: Seamless speech interaction with large language models.
\newblock \emph{arXiv preprint arXiv:2409.06666}, 2024.

\bibitem[Fang et~al.(2025)Fang, Zhou, Guo, Zhang, and Feng]{fang2025llama}
Qingkai Fang, Yan Zhou, Shoutao Guo, Shaolei Zhang, and Yang Feng.
\newblock Llama-omni2: Llm-based real-time spoken chatbot with autoregressive streaming speech synthesis.
\newblock \emph{arXiv preprint arXiv:2505.02625}, 2025.

\bibitem[Fu et~al.(2023)Fu, Chen, Shen, Qin, Zhang, Lin, Yang, Zheng, Li, Sun, et~al.]{fu2023mme}
Chaoyou Fu, Peixian Chen, Yunhang Shen, Yulei Qin, Mengdan Zhang, Xu~Lin, Jinrui Yang, Xiawu Zheng, Ke~Li, Xing Sun, et~al.
\newblock Mme: A comprehensive evaluation benchmark for multimodal large language models.
\newblock \emph{arXiv preprint arXiv:2306.13394}, 2023.

\bibitem[Fu et~al.(2024{\natexlab{a}})Fu, Dai, Luo, Li, Ren, Zhang, Wang, Zhou, Shen, Zhang, et~al.]{fu2024video}
Chaoyou Fu, Yuhan Dai, Yongdong Luo, Lei Li, Shuhuai Ren, Renrui Zhang, Zihan Wang, Chenyu Zhou, Yunhang Shen, Mengdan Zhang, et~al.
\newblock Video-mme: The first-ever comprehensive evaluation benchmark of multi-modal llms in video analysis.
\newblock \emph{arXiv preprint arXiv:2405.21075}, 2024{\natexlab{a}}.

\bibitem[Fu et~al.(2024{\natexlab{b}})Fu, Lin, Long, Shen, Zhao, Zhang, Dong, Wang, Yin, Ma, et~al.]{fu2024vita}
Chaoyou Fu, Haojia Lin, Zuwei Long, Yunhang Shen, Meng Zhao, Yifan Zhang, Shaoqi Dong, Xiong Wang, Di~Yin, Long Ma, et~al.
\newblock Vita: Towards open-source interactive omni multimodal llm.
\newblock \emph{arXiv preprint arXiv:2408.05211}, 2024{\natexlab{b}}.

\bibitem[Grattafiori et~al.(2024)]{meta2024llama3}
Aaron Grattafiori et~al.
\newblock The llama 3 herd of models.
\newblock \emph{IEEE Spectrum}, 2024.

\bibitem[Guo et~al.(2025)Guo, Song, Feng, Ma, Zhang, Gao, Yu, Sun, Chen, Yang, et~al.]{guo2025m2}
Qingpei Guo, Kaiyou Song, Zipeng Feng, Ziping Ma, Qinglong Zhang, Sirui Gao, Xuzheng Yu, Yunxiao Sun, Jingdong Chen, Ming Yang, et~al.
\newblock M2-omni: Advancing omni-mllm for comprehensive modality support with competitive performance.
\newblock \emph{arXiv preprint arXiv:2502.18778}, 2025.

\bibitem[Gupta and Kembhavi(2023)]{gupta2023visual}
Tanmay Gupta and Aniruddha Kembhavi.
\newblock Visual programming: Compositional visual reasoning without training.
\newblock In \emph{Proceedings of the IEEE/CVF Conference on Computer Vision and Pattern Recognition}, pages 14953--14962, 2023.

\bibitem[Hong et~al.(2025)Hong, Yan, Cai, Jiang, Hu, and Xie]{hong2025worldsense}
Jack Hong, Shilin Yan, Jiayin Cai, Xiaolong Jiang, Yao Hu, and Weidi Xie.
\newblock Worldsense: Evaluating real-world omnimodal understanding for multimodal llms.
\newblock \emph{arXiv preprint arXiv:2502.04326}, 2025.

\bibitem[Hurst et~al.(2024)Hurst, Lerer, Goucher, Perelman, Ramesh, Clark, Ostrow, Welihinda, Hayes, Radford, et~al.]{hurst2024gpt}
Aaron Hurst, Adam Lerer, Adam~P Goucher, Adam Perelman, Aditya Ramesh, Aidan Clark, AJ~Ostrow, Akila Welihinda, Alan Hayes, Alec Radford, et~al.
\newblock Gpt-4o system card.
\newblock \emph{arXiv preprint arXiv:2410.21276}, 2024.

\bibitem[Ishibashi and Nishimura(2024)]{ishibashi2024self}
Yoichi Ishibashi and Yoshimasa Nishimura.
\newblock Self-organized agents: A llm multi-agent framework toward ultra large-scale code generation and optimization.
\newblock \emph{arXiv preprint arXiv:2404.02183}, 2024.

\bibitem[Kumar et~al.(2024)Kumar, Gadhia, Ganu, and Nambi]{kumar2024mmctagent}
Somnath Kumar, Yash Gadhia, Tanuja Ganu, and Akshay Nambi.
\newblock Mmctagent: Multi-modal critical thinking agent framework for complex visual reasoning.
\newblock \emph{arXiv preprint arXiv:2405.18358}, 2024.

\bibitem[Liang et~al.(2024)Liang, Tao, Xia, Shi, Wang, and Yang]{liang2024cmat}
Xuechen Liang, Meiling Tao, Yinghui Xia, Tianyu Shi, Jun Wang, and JingSong Yang.
\newblock Cmat: A multi-agent collaboration tuning framework for enhancing small language models.
\newblock \emph{arXiv preprint arXiv:2404.01663}, 2024.

\bibitem[Liu et~al.(2024{\natexlab{a}})Liu, Li, Li, Li, Zhang, Shen, and Lee]{liu2024llavanext}
Haotian Liu, Chunyuan Li, Yuheng Li, Bo~Li, Yuanhan Zhang, Sheng Shen, and Yong~Jae Lee.
\newblock {LLaVA-NeXT}: Improved reasoning, {OCR}, and world knowledge, January 2024{\natexlab{a}}.
\newblock URL \url{https://llava-vl.github.io/blog/2024-01-30-llava-next/}.

\bibitem[Liu et~al.(2023)]{liu2023llava}
Haotian Liu et~al.
\newblock Visual instruction tuning.
\newblock \emph{arXiv preprint arXiv:2304.08485}, 2023.

\bibitem[Liu et~al.(2024{\natexlab{b}})Liu, Duan, Zhang, Li, Zhang, Zhao, Yuan, Wang, He, Liu, et~al.]{liu2024mmbench}
Yuan Liu, Haodong Duan, Yuanhan Zhang, Bo~Li, Songyang Zhang, Wangbo Zhao, Yike Yuan, Jiaqi Wang, Conghui He, Ziwei Liu, et~al.
\newblock Mmbench: Is your multi-modal model an all-around player?
\newblock In \emph{European conference on computer vision}, pages 216--233. Springer, 2024{\natexlab{b}}.

\bibitem[Lu et~al.(2025)Lu, Li, Cong, Zhang, Wu, Lin, Liu, Liu, and Sun]{lu2025learning}
Yaxi Lu, Haolun Li, Xin Cong, Zhong Zhang, Yesai Wu, Yankai Lin, Zhiyuan Liu, Fangming Liu, and Maosong Sun.
\newblock Learning to generate structured output with schema reinforcement learning.
\newblock \emph{arXiv preprint arXiv:2502.18878}, 2025.

\bibitem[OpenAI et~al.(2023)]{openai2023gpt4}
OpenAI et~al.
\newblock Gpt-4 technical report.
\newblock Technical report, OpenAI, 2023.

\bibitem[Qiao et~al.(2023)Qiao, Li, Zhang, He, Kang, Zhang, Yang, Dong, Zhang, Wang, et~al.]{qiao2023taskweaver}
Bo~Qiao, Liqun Li, Xu~Zhang, Shilin He, Yu~Kang, Chaoyun Zhang, Fangkai Yang, Hang Dong, Jue Zhang, Lu~Wang, et~al.
\newblock Taskweaver: A code-first agent framework.
\newblock \emph{arXiv preprint arXiv:2311.17541}, 2023.

\bibitem[Shen et~al.(2023)Shen, Song, Tan, Li, Lu, and Zhuang]{shen2023hugginggpt}
Yongliang Shen, Kaitao Song, Xu~Tan, Dongsheng Li, Weiming Lu, and Yueting Zhuang.
\newblock Hugginggpt: Solving ai tasks with chatgpt and its friends in hugging face.
\newblock \emph{arXiv preprint arXiv:2303.17580}, 2023.

\bibitem[Shin et~al.(2024)Shin, Lim, Won, Choi, Kim, Song, Yoo, Kim, and Lim]{shin2024x}
DongJae Shin, HyeonSeok Lim, Inho Won, ChangSu Choi, Minjun Kim, SeungWoo Song, HanGyeol Yoo, SangMin Kim, and KyungTae Lim.
\newblock {X-LLaVA: Optimizing Bilingual Large Vision-Language Alignment}.
\newblock In \emph{Findings of the Association for Computational Linguistics: NAACL 2024}, pages 2463--2473, Mexico City, Mexico, June 2024. Association for Computational Linguistics.
\newblock \doi{10.18653/v1/2024.findings-naacl.158}.
\newblock URL \url{https://aclanthology.org/2024.findings-naacl.158}.

\bibitem[Sun et~al.(2024)Sun, Dai, Luo, Chang, and Li]{sun2024lawluo}
Jingyun Sun, Chengxiao Dai, Zhongze Luo, Yangbo Chang, and Yang Li.
\newblock {LawLuo: A Multi-Agent Collaborative Framework for Multi-Round Chinese Legal Consultation}.
\newblock \emph{arXiv preprint arXiv:2407.16252v3}, 2024.
\newblock \doi{10.48550/arXiv.2407.16252}.

\bibitem[Sur{\'\i}s et~al.(2023)Sur{\'\i}s, Menon, and Vondrick]{suris2023vipergpt}
D{\'\i}dac Sur{\'\i}s, Sachit Menon, and Carl Vondrick.
\newblock Vipergpt: Visual inference via python execution for reasoning.
\newblock In \emph{Proceedings of the IEEE/CVF International Conference on Computer Vision}, pages 11888--11898, 2023.

\bibitem[Team et~al.(2024)Team, Georgiev, Lei, Burnell, Bai, Gulati, Tanzer, Vincent, Pan, Wang, et~al.]{team2024gemini}
Gemini Team, Petko Georgiev, Ving~Ian Lei, Ryan Burnell, Libin Bai, Anmol Gulati, Garrett Tanzer, Damien Vincent, Zhufeng Pan, Shibo Wang, et~al.
\newblock Gemini 1.5: Unlocking multimodal understanding across millions of tokens of context.
\newblock \emph{arXiv preprint arXiv:2403.05530}, 2024.

\bibitem[Team et~al.(2023)]{gemini2023}
Gemini Team et~al.
\newblock Gemini: A family of highly capable multimodal models.
\newblock \emph{arXiv preprint arXiv:2312.11805}, 2023.

\bibitem[Tong et~al.(2024)Tong, Brown, Wu, Woo, IYER, Akula, Yang, Yang, Middepogu, Wang, et~al.]{tong2024cambrian}
Peter Tong, Ellis Brown, Penghao Wu, Sanghyun Woo, Adithya Jairam~Vedagiri IYER, Sai~Charitha Akula, Shusheng Yang, Jihan Yang, Manoj Middepogu, Ziteng Wang, et~al.
\newblock Cambrian-1: A fully open, vision-centric exploration of multimodal llms.
\newblock \emph{Advances in Neural Information Processing Systems}, 37:\penalty0 87310--87356, 2024.

\bibitem[Wang et~al.(2025)Wang, Luo, Dong, Xuan, Li, Ma, and Gao]{wang2025mllm}
Chenyu Wang, Weixin Luo, Sixun Dong, Xiaohua Xuan, Zhengxin Li, Lin Ma, and Shenghua Gao.
\newblock Mllm-tool: A multimodal large language model for tool agent learning.
\newblock In \emph{2025 IEEE/CVF Winter Conference on Applications of Computer Vision (WACV)}, pages 6678--6687. IEEE, 2025.

\bibitem[Wang et~al.(2024{\natexlab{a}})Wang, Pan, Shi, Lu, Ren, Zhou, Zhan, and Li]{wang2024measuring}
Ke~Wang, Junting Pan, Weikang Shi, Zimu Lu, Houxing Ren, Aojun Zhou, Mingjie Zhan, and Hongsheng Li.
\newblock Measuring multimodal mathematical reasoning with math-vision dataset.
\newblock \emph{Advances in Neural Information Processing Systems}, 37:\penalty0 95095--95169, 2024{\natexlab{a}}.

\bibitem[Wang et~al.(2024{\natexlab{b}})Wang, Bai, Tan, Wang, Fan, Bai, Chen, Liu, Wang, Ge, et~al.]{wang2024qwen2}
Peng Wang, Shuai Bai, Sinan Tan, Shijie Wang, Zhihao Fan, Jinze Bai, Keqin Chen, Xuejing Liu, Jialin Wang, Wenbin Ge, et~al.
\newblock Qwen2-vl: Enhancing vision-language model's perception of the world at any resolution.
\newblock \emph{arXiv preprint arXiv:2409.12191}, 2024{\natexlab{b}}.

\bibitem[Wang et~al.(2024{\natexlab{c}})Wang, He, Hong, Cheng, Zhang, Qi, Gu, Huang, Xu, Dong, et~al.]{wang2024lvbench}
Weihan Wang, Zehai He, Wenyi Hong, Yean Cheng, Xiaohan Zhang, Ji~Qi, Xiaotao Gu, Shiyu Huang, Bin Xu, Yuxiao Dong, et~al.
\newblock Lvbench: An extreme long video understanding benchmark.
\newblock \emph{arXiv preprint arXiv:2406.08035}, 2024{\natexlab{c}}.

\bibitem[Wang et~al.(2024{\natexlab{d}})Wang, Li, Fu, Shen, Xie, Li, Sun, and Ma]{wang2024freeze}
Xiong Wang, Yangze Li, Chaoyou Fu, Yunhang Shen, Lei Xie, Ke~Li, Xing Sun, and Long Ma.
\newblock Freeze-omni: A smart and low latency speech-to-speech dialogue model with frozen llm.
\newblock \emph{arXiv preprint arXiv:2411.00774}, 2024{\natexlab{d}}.

\bibitem[Wu et~al.(2023)Wu, Bansal, Zhang, Wu, Li, Zhu, Jiang, Zhang, Zhang, Liu, et~al.]{wu2023autogen}
Qingyun Wu, Gagan Bansal, Jieyu Zhang, Yiran Wu, Beibin Li, Erkang Zhu, Li~Jiang, Xiaoyun Zhang, Shaokun Zhang, Jiale Liu, et~al.
\newblock Autogen: Enabling next-gen llm applications via multi-agent conversation.
\newblock \emph{arXiv preprint arXiv:2308.08155}, 2023.

\bibitem[Xie and Wu(2024)]{xie2024mini}
Zhifei Xie and Changqiao Wu.
\newblock Mini-omni2: Towards open-source gpt-4o with vision, speech and duplex capabilities.
\newblock \emph{arXiv preprint arXiv:2410.11190}, 2024.

\bibitem[Xu et~al.(2025{\natexlab{a}})Xu, Guo, He, Hu, He, Bai, Chen, Wang, Fan, Dang, Zhang, Wang, Chu, and Lin]{Qwen2.5-Omni}
Jin Xu, Zhifang Guo, Jinzheng He, Hangrui Hu, Ting He, Shuai Bai, Keqin Chen, Jialin Wang, Yang Fan, Kai Dang, Bin Zhang, Xiong Wang, Yunfei Chu, and Junyang Lin.
\newblock {Qwen2.5-Omni Technical Report}.
\newblock \emph{arXiv preprint arXiv:2503.20215}, 2025{\natexlab{a}}.
\newblock \doi{10.48550/arXiv.2503.20215}.

\bibitem[Xu et~al.(2025{\natexlab{b}})Xu, Liang, Mei, Gao, Tan, and Zhang]{xu2025mem}
Wujiang Xu, Zujie Liang, Kai Mei, Hang Gao, Juntao Tan, and Yongfeng Zhang.
\newblock A-mem: Agentic memory for llm agents.
\newblock \emph{arXiv preprint arXiv:2502.12110}, 2025{\natexlab{b}}.

\bibitem[Yang et~al.(2025)Yang, Li, Yang, Zhang, Hui, Zheng, Yu, Gao, Huang, Lv, et~al.]{yang2025qwen3}
An~Yang, Anfeng Li, Baosong Yang, Beichen Zhang, Binyuan Hui, Bo~Zheng, Bowen Yu, Chang Gao, Chengen Huang, Chenxu Lv, et~al.
\newblock Qwen3 technical report.
\newblock \emph{arXiv preprint arXiv:2505.09388}, 2025.

\bibitem[Yang et~al.(2024)Yang, Tang, Li, Wang, Wan, Zhong, Liu, Yang, Wang, Liu, et~al.]{yang2024cc}
Zhibo Yang, Jun Tang, Zhaohai Li, Pengfei Wang, Jianqiang Wan, Humen Zhong, Xuejing Liu, Mingkun Yang, Peng Wang, Yuliang Liu, et~al.
\newblock Cc-ocr: A comprehensive and challenging ocr benchmark for evaluating large multimodal models in literacy.
\newblock \emph{arXiv preprint arXiv:2412.02210}, 2024.

\bibitem[Yao et~al.(2023)Yao, Zhao, Yu, Du, Shafran, Narasimhan, and Cao]{yao2023react}
Shunyu Yao, Jeffrey Zhao, Dian Yu, Nan Du, Izhak Shafran, Karthik Narasimhan, and Yuan Cao.
\newblock {ReAct}: Synergizing reasoning and acting in language models.
\newblock In \emph{International Conference on Learning Representations (ICLR)}, 2023.

\bibitem[Yu et~al.(2024)Yu, Wang, Yang, Chen, Tian, Zhang, Sun, Lu, Wang, and Zhang]{yu2024salmonn}
Wenyi Yu, Siyin Wang, Xiaoyu Yang, Xianzhao Chen, Xiaohai Tian, Jun Zhang, Guangzhi Sun, Lu~Lu, Yuxuan Wang, and Chao Zhang.
\newblock Salmonn-omni: A codec-free llm for full-duplex speech understanding and generation.
\newblock \emph{arXiv preprint arXiv:2411.18138}, 2024.

\bibitem[Yue et~al.(2024)Yue, Ni, Zhang, Zheng, Liu, Zhang, Stevens, Jiang, Ren, Sun, et~al.]{yue2024mmmu}
Xiang Yue, Yuansheng Ni, Kai Zhang, Tianyu Zheng, Ruoqi Liu, Ge~Zhang, Samuel Stevens, Dongfu Jiang, Weiming Ren, Yuxuan Sun, et~al.
\newblock Mmmu: A massive multi-discipline multimodal understanding and reasoning benchmark for expert agi.
\newblock In \emph{Proceedings of the IEEE/CVF Conference on Computer Vision and Pattern Recognition}, pages 9556--9567, 2024.

\bibitem[Zeng et~al.(2024)Zeng, Fang, Liu, and Meng]{zeng2024structural}
Ruihong Zeng, Jinyuan Fang, Siwei Liu, and Zaiqiao Meng.
\newblock On the structural memory of llm agents.
\newblock \emph{arXiv preprint arXiv:2412.15266}, 2024.

\bibitem[Zhang et~al.(2023)Zhang, Dong, Wang, Cao, Xu, Ouyang, Zhao, Duan, Zhang, Ding, et~al.]{zhang2023internlm}
Pan Zhang, Xiaoyi Dong, Bin Wang, Yuhang Cao, Chao Xu, Linke Ouyang, Zhiyuan Zhao, Haodong Duan, Songyang Zhang, Shuangrui Ding, et~al.
\newblock Internlm-xcomposer: A vision-language large model for advanced text-image comprehension and composition.
\newblock \emph{arXiv preprint arXiv:2309.15112}, 2023.

\bibitem[Zhao et~al.(2024)Zhao, Yang, Peng, Bai, Yao, Sun, Chen, Fu, Chen, Wei, and Bo]{zhao2024humanomni}
Jiaxing Zhao, Qize Yang, Yixing Peng, Detao Bai, Shimin Yao, Boyuan Sun, Xiang Chen, Shenghao Fu, Weixuan Chen, Xihan Wei, and Liefeng Bo.
\newblock Humanomni: A large vision-speech language model for human-centric video understanding.
\newblock \emph{arXiv preprint arXiv:2501.15111}, 2024.

\bibitem[Zhu et~al.(2023)Zhu, Chen, Shen, Li, and Elhoseiny]{zhu2023minigpt}
Deyao Zhu, Jun Chen, Xiaoqian Shen, Xiang Li, and Mohamed Elhoseiny.
\newblock Minigpt-4: Enhancing vision-language understanding with advanced large language models.
\newblock \emph{arXiv preprint arXiv:2304.10592}, 2023.

\end{thebibliography}

\clearpage
\beginappendix
\small
\setlength{\parindent}{1em}
\setlength{\parskip}{0.35em}
\setlength{\textfloatsep}{10pt}
\setlength{\floatsep}{8pt}
\setlength{\intextsep}{8pt}
\setlength{\abovecaptionskip}{4pt}
\setlength{\belowcaptionskip}{2pt}
\section*{Appendix Roadmap}
\addcontentsline{toc}{section}{Appendix}
\vspace{-0.4em}
\begin{center}
\small
\setlength{\tabcolsep}{0pt}
\begin{tabular*}{0.96\textwidth}{@{\extracolsep{\fill}} l r}
\textbf{A}~~Core Instruction Set: Communication Protocol & \pageref{sec:instruction_set} \\
\textbf{B}~~Core Prompt Design: Dialogue Control Framework & \pageref{sec:prompt_design} \\
\textbf{C}~~Memory Pool Schema: Cross-modal Memory Management & \pageref{sec:memory_json} \\
\textbf{D}~~Modality-Specific Processing Details & \pageref{sec:tts_rules} \\
\textbf{E}~~LLM Orchestration Algorithm & \pageref{sec:algorithm} \\
\textbf{F}~~Additional Experimental Results & \pageref{sec:appendix_additional_results} \\
\end{tabular*}
\end{center}
\vspace{0.6em}

\section{Core Instruction Set: Communication Protocol}
\label{sec:instruction_set}

The controller and router communicate through a closed-vocabulary token protocol. Each routing token is bound to a mapping entry that specifies the target expert, required inputs, output schema, and runtime policies (e.g., timeout, cancellability). Interaction tokens are handled by the interaction manager and do not invoke experts.

\begin{table}[t]
\centering
\renewcommand{\arraystretch}{1.10}
\scriptsize
\setlength{\tabcolsep}{1.2mm}
\caption{Control token vocabulary used in all experiments. Routing tokens trigger mapping-bound expert calls; interaction tokens affect dialogue state and execution control. The pattern \emph{[S.need\_*]} is extensible via the mapping without changing the parser or router.}
\label{tab:control_tokens}
\begin{tabular}{@{}p{0.26\columnwidth}p{0.13\columnwidth}p{0.52\columnwidth}@{}}
\toprule
\textbf{Token} & \textbf{Type} & \textbf{Router semantics} \\
\midrule
\multicolumn{3}{@{}l}{\textit{Routing tokens}} \\
\addlinespace[0.2em]
\texttt{[S.need\_vision]}    & route & invoke vision expert on referenced media segment(s) \\
\texttt{[S.need\_video]}     & route & invoke video expert on sampled frames / video segment(s) \\
\texttt{[S.need\_audio]}     & route & invoke ASR/audio expert and return transcript/evidence \\
\texttt{[S.need\_ocr]}       & route & invoke OCR expert when text extraction is required \\
\texttt{[S.need\_math]}      & route & invoke math specialist when required by benchmark \\
\addlinespace[0.4em]
\multicolumn{3}{@{}l}{\textit{Interaction tokens (dialogue control)}} \\
\addlinespace[0.2em]
\texttt{[S.listen]}          & interact & pause and await additional user input; keep state \\
\texttt{[S.stop]}            & interact & cancel in-flight execution when permitted by policy \\
\texttt{[S.speak]}           & interact & allow downstream TTS rendering of the final text response \\
\bottomrule
\end{tabular}
\end{table}

\noindent\textbf{Deterministic validation.}
The router validates that the controller output is tokens-only and that all routing tokens are well-formed and mapping-resolvable. Invalid outputs trigger rejection and a safe fallback route that produces a bounded text-only response.

\section{Core Prompt Design: Dialogue Control Framework}
\label{sec:prompt_design}

We provide the controller with (i) a protocol specification defining the closed token vocabulary, (ii) a mapping describing expert interfaces and runtime policies, and (iii) routing guidelines that encourage evidence reuse when valid keys are available. The generator receives only text evidence (retrieved expert outputs and compact memory view) and is never parsed for control tokens.

\noindent\textbf{Controller prompt skeleton.}
\begin{tcolorbox}[appendixcodebox]
\raggedright
\textbf{Role:} You are the orchestration controller.\\
\textbf{Input:} (a) user query $q_t$; (b) compact memory view $\widehat{M}_{t-1}$; (c) mapping summary.\\
\textbf{Output constraint:} Emit a sequence of control tokens only, each from the allowed vocabulary. Do not output any other text.\\
\textbf{Routing guideline:} Prefer reusing cached evidence when a matching evidence key exists; avoid redundant expert calls.\\
\textbf{Failure mode:} Invalid outputs will be rejected by a strict parser and replaced by a safe default route.
\end{tcolorbox}

\noindent\textbf{Mapping snippet (illustrative).}
\begin{tcolorbox}[appendixcodebox]
\textbf{Token} \texttt{[S.need\_vision]} $\rightarrow$ \textbf{Expert:} \texttt{vision\_expert}\\
\textbf{Inputs:} current image/video frames OR reusable \texttt{evidence\_key}\\
\textbf{Output schema:} structured textual evidence (caption/objects/attributes)\\
\textbf{Policy:} timeout $T_{\text{vision}}$, cancellable=true
\par\medskip
\textbf{Token} \texttt{[S.need\_audio]} $\rightarrow$ \textbf{Expert:} \texttt{asr\_expert}\\
\textbf{Inputs:} current audio segment OR reusable \texttt{evidence\_key}\\
\textbf{Output schema:} transcript + timestamps\\
\textbf{Policy:} timeout $T_{\text{asr}}$, cancellable=true
\end{tcolorbox}

\noindent\textbf{Generator prompting.}
The generator is invoked with a disjoint prompt that contains $(q_t,\widehat{M}_{t-1},D_t)$ and is required to produce a final text response only. When the same underlying LLM is used for both roles, we use two separate calls with non-overlapping prompts; only the controller call is parsed for control tokens.

\section{Memory Pool Schema: Cross-modal Memory Management}
\label{sec:memory_json}

The memory pool stores expert outputs as structured text records indexed by evidence keys. The goal is verifiable evidence reuse under bounded latency, rather than long-horizon persona storage. Persistent writes are restricted to completed expert outputs and finalized segments; partial hypotheses and canceled calls remain ephemeral (Appendix~\ref{sec:tts_rules}).

\noindent\textbf{Record format.}
Each memory entry is a tuple:
\[
e_j=\{\text{turn},\text{modality},\text{evidence\_key},\text{payload},\text{meta}\},
\]
where \emph{payload} is a compact textual/JSON-like expert output, and \emph{evidence\_key} identifies the originating input segment.

\begin{tcolorbox}[appendixcodebox,title=Memory entry examples (illustrative)]
\scriptsize
\# Evidence key\\
\texttt{key} := (\texttt{modality}, \texttt{source}, \texttt{segment}, \texttt{checksum})\\
\par\medskip
\# Vision entry (abridged)\\
\{\\
~~"turn": 3,\\
~~"modality": "vision",\\
~~"key": ["vision","img\_042","full","sha1:..."],\\
~~"payload": \{"caption": "..."\},\\
~~"meta": \{"model": "Qwen2.5-VL-32B"\}\\
\}\\
\par\medskip
\# ASR entry (abridged)\\
\{\\
~~"turn": 4,\\
~~"modality": "audio",\\
~~"key": ["audio","aud\_017","t=12-18","sha1:..."],\\
~~"payload": \{"transcript": "..."\},\\
~~"meta": \{"model": "ASR"\}\\
\}
\end{tcolorbox}

\noindent\textbf{Compression view.}
To respect context-length limits, we maintain a compact view $\widehat{M}_t$ by selecting a bounded set of records using a weighted score that combines recency, lexical relevance to $q_t$, and modality coverage. The full memory pool $M_t$ is retained for router-side retrieval via evidence keys; the controller and generator read only $\widehat{M}_t$.

\section{Modality-Specific Processing Details}
\label{sec:tts_rules}

This section documents implementation details for standard input preprocessing and output rendering modules that operate outside the core orchestration logic. These modules handle modality-specific transformations to ensure compatibility with the controller's text-centric interface.

\noindent\textbf{Input Preprocessing:} Raw multimodal inputs are converted to text-compatible formats via standard modules: ASR (Automatic Speech Recognition) for audio input, OCR (Optical Character Recognition) for text extraction from images, and frame extraction for video preprocessing. These preprocessing steps are orthogonal to orchestration and utilize off-the-shelf tools.

\noindent\textbf{Output Rendering - TTS Segmentation:} For interactive voice applications, the text response $o_t$ may be synthesized to speech via TTS (Text-to-Speech). The TTS module employs a rule-based segmentation strategy to optimize speech synthesis quality and latency. The segmentation rules are designed to maintain semantic coherence while enabling parallel processing. Below are the detailed segmentation rules:

\begin{tcolorbox}[appendixbox,title=TTS Segmentation Rules]
\emph{Rule 1}: Natural punctuation boundaries — e.g., periods, commas, semicolons

\emph{Rule 2}: Discourse markers and conjunctions — e.g., "however", "therefore", "and"

\emph{Rule 3}: Syntactic boundaries — e.g., subject-predicate splits, subordinate clauses
\end{tcolorbox}

Each segment typically contains 7-15 words to balance synthesis quality and parallelization efficiency. This granularity ensures both semantic completeness and natural prosody while maintaining computational efficiency. For example, the sentence "The cat sat on the mat, while the dog slept peacefully" would be split into two segments: "The cat sat on the mat" and "while the dog slept peacefully".

\noindent\textbf{Commit rules.}
Canceled TTS segments are not written into memory. For audio inputs, only finalized ASR transcripts are committed for reuse; partial hypotheses remain ephemeral.

\section{LLM Orchestration Algorithm}
\label{sec:algorithm}

This section provides the full orchestration pipeline used throughout our experiments. The controller emits a \emph{tokens-only} plan under a closed vocabulary; the router deterministically validates and dispatches mapping-bound actions with cache-or-call semantics; a text-only generator produces the final response conditioned on retrieved evidence. Modality-specific preprocessing (ASR/OCR/frame extraction) and rendering (TTS) are standard modules and are detailed in Appendix~\ref{sec:tts_rules}.

\begin{algorithm}[H]
    \scriptsize
    \caption{Training-free LLM Orchestration (Controller--Router--Generator)}
    \label{alg:dynamic_orchestration}
    \begin{algorithmic}[1]
    \STATE {\bfseries Input:} Multimodal user input $u_t$, text query $q_t$, compact memory view $\widehat{M}_{t-1}$, full memory pool $M_{t-1}$, expert mapping $\mathcal{E}$
    \STATE {\bfseries Output:} Text response $o_t$ (optionally synthesized to audio by TTS)
    \STATE \textbf{procedure} \textsc{Orchestrate}($u_t,q_t,\widehat{M}_{t-1},M_{t-1},\mathcal{E}$)
    \STATE \quad $S_t \gets f_{\text{ctrl}}(q_t,\widehat{M}_{t-1})$ \hfill // controller emits \emph{control tokens only}
    \STATE \quad \textbf{if} $\neg\,\textsc{IsValidTokensOnly}(S_t)$ \textbf{then}
    \STATE \qquad $S_t \gets \textsc{SafeDefaultRoute}()$ \hfill // no external expert calls
    \STATE \quad \textbf{end if}
    \STATE \quad \textbf{if} $\texttt{[S.stop]} \in S_t$ \textbf{then} \textbf{return} \textit{null} \hfill // interruption handled by interaction manager
    \STATE \quad \textbf{if} $\texttt{[S.listen]} \in S_t$ \textbf{then} \textbf{return} \textit{null} \hfill // await additional input
    \STATE \quad $\delta(S_t) \gets \textsc{ValidateAndPlan}(S_t,\mathcal{E})$ \hfill // deterministic parsing + mapping checks
    \STATE \quad $D_t \gets \textsc{CacheOrCall}(\delta(S_t),u_t,q_t,M_{t-1})$ \hfill // hit-or-miss per action
    \STATE \quad $z_t \gets \textsc{Compose}(q_t,\widehat{M}_{t-1},D_t)$ \hfill // text-only evidence context
    \STATE \quad $o_t \gets f_{\text{gen}}(z_t)$ \hfill // generator emits \emph{text only}
    \STATE \quad $M_t \gets \textsc{Commit}(M_{t-1},q_t,D_t,o_t)$ \hfill // commit only completed outputs
    \STATE \quad $\widehat{M}_t \gets h_{\text{compress}}(M_t)$ \hfill // bounded memory view for next turn
    \STATE \quad \textbf{return} $o_t$
    \STATE \textbf{end procedure}
    \STATE \textbf{function} \textsc{ValidateAndPlan}($S_t,\mathcal{E}$)
    \STATE \quad parse $S_t$ and keep routing tokens $s_k \in \mathcal{S}_{\text{route}}$
    \STATE \quad map each routing token to a mapping-bound action $a_k=\Gamma(s_k)$
    \STATE \quad \textbf{return} ordered plan $\delta(S_t)=[a_1,\ldots,a_{K'}]$
    \STATE \textbf{end function}
    \STATE \textbf{function} \textsc{CacheOrCall}($\delta(S_t),u_t,q_t,M_{t-1}$)
    \STATE \quad $D_t \gets [\,]$
    \FOR{$a_k$ \textbf{in} $\delta(S_t)$}
        \STATE \quad $(\texttt{hit}, d) \gets \textsc{Retrieve}(a_k,u_t,q_t,M_{t-1})$
        \STATE \quad \textbf{if} $\neg\,\texttt{hit}$ \textbf{then} $d \gets \textsc{RunExpert}(\pi(a_k),a_k,u_t)$ \hfill // bounded by runtime policy
        \STATE \quad append $d$ to $D_t$
    \ENDFOR
    \STATE \quad \textbf{return} $D_t$
    \STATE \textbf{end function}
    \end{algorithmic}
\end{algorithm}

\paragraph{Multi-turn Full-Duplex Workflow.}
We include the full multi-turn interaction loop used in the main paper.
\begin{algorithm}[H]
\scriptsize
\caption{Multi-turn Full-Duplex Workflow}
\label{alg:multiturn-workflow}
\begin{algorithmic}[1]
\REQUIRE Mapping $\mathcal{E}$ with policies (timeout, cancellable); protocol vocabulary $\mathcal{S}$
\STATE $M\leftarrow \emptyset$; $\hat{M}\leftarrow \emptyset$; playback buffer $\mathcal{B}\leftarrow \emptyset$; async jobs $\mathcal{J}\leftarrow \emptyset$
\WHILE{session active}
  \STATE $\tilde{u}\leftarrow \textsc{Stage1}(u)$; $q\leftarrow \textsc{SelectQuery}(\tilde{u})$; $\mathcal{K}\leftarrow \textsc{Keys}(\tilde{u})$
  \IF{\textsc{BargeInDetected}($\tilde{u}$) \textbf{and} $\mathcal{B}\neq \emptyset$}
    \STATE \textsc{CancelInFlightJobs}($\mathcal{J},\mathcal{E}$); $\mathcal{B}\leftarrow \emptyset$
  \ENDIF
  \STATE $S\leftarrow f_{\text{ctrl}}(q,\hat{M})$
  \STATE $\delta\leftarrow \textsc{ValidateAndPlan}(S;\mathcal{E},\mathcal{K})$
  \STATE \textbf{parallel for} $a\in\delta$ \textbf{do} $d(a)\leftarrow \textsc{CacheOrCall}(a,\tilde{u},M;\mathcal{E})$; \textsc{Enqueue}($\mathcal{J}$, job$(a)$) \textbf{end for}
  \STATE $D\leftarrow [d(a)\mid a\in\delta]$; $o^{\text{text}}\leftarrow f_{\text{gen}}(\textsc{Compose}(q,\hat{M},D))$
  \STATE $M\leftarrow \textsc{Commit}(M,D,o^{\text{text}})$; $\hat{M}\leftarrow \textsc{Compress}(M)$
  \IF{\textsc{SpeakEnabled}($S$)}
    \STATE $\{c_i\}_{i=1}^{B}\leftarrow \textsc{Chunk}(o^{\text{text}};\mathcal{P},L)$
    \STATE \textbf{parallel for} $i=1$ to $B$ \textbf{do} $J_i\leftarrow \textsc{AsyncTTS}(c_i)$; \textsc{Enqueue}($\mathcal{J}$, $J_i$) \textbf{end for}
    \STATE \textbf{for} $i=1$ to $B$ \textbf{do}
      \IF{\textsc{StopFlagRaised}() \textbf{or} \textsc{BargeInDetected}(\textsc{Stage1}(u))}
        \STATE \textsc{CancelPendingTTS}($\{J_j\}_{j\ge i}$); \textbf{break}
      \ENDIF
      \STATE $y_i\leftarrow \textsc{Await}(J_i)$; \textsc{StreamPlay}($y_i$); update $\mathcal{B}$
    \STATE \textbf{end for}
  \ELSE
    \STATE \textsc{EmitText}($o^{\text{text}}$)
  \ENDIF
\ENDWHILE
\end{algorithmic}
\end{algorithm}

\section{Additional Experimental Results}
\label{sec:appendix_additional_results}

\subsection{Parameter-Matched Routing Analysis}
\label{sec:appendix_param_matched}

To separate the benefit of orchestration from raw expert scale, we evaluate routed configurations with different average active visual-parameter budgets on Video-MME. The controller is fixed, while the visual expert pool contains Qwen2.5-VL-7B, Qwen2.5-VL-32B, and Qwen2.5-VL-72B. Each routed variant invokes a mixture of experts; therefore, \emph{Avg. Visual Expert Params} denotes the weighted average parameter count of invoked visual experts rather than the maximum available expert size. The main table reports the best routed setting as 53.8B average active visual expert parameters; the 14B controller is fixed across routed variants.

\begin{table}[H]
  \centering
  \caption{Parameter-matched Video-MME analysis under a fixed 14B controller. Routed variants differ only in the distribution over the visual expert pool.}
  \label{tab:param_matched_routing}
  \renewcommand{\arraystretch}{1.06}
  \scriptsize
  \setlength{\tabcolsep}{0.45mm}
  \begin{tabular*}{\columnwidth}{@{\extracolsep{\fill}}lccccc@{}}
    \toprule
    \textbf{Model} & \textbf{Avg.} & \textbf{Acc.} & \textbf{7B} & \textbf{32B} & \textbf{72B} \\
    & \textbf{(B)} & \textbf{(\%)} & \textbf{(\%)} & \textbf{(\%)} & \textbf{(\%)} \\
    \midrule
    Qwen2.5-VL & 72.0 & 65.74 & -- & -- & -- \\
    LLaVA-OV & 72.0 & 66.20 & -- & -- & -- \\
    Ours (Qwen2.5-VL) & 19.7 & 62.07 & 49.20 & 50.80 & 0.00 \\
    Ours (Qwen2.5-VL) & 35.2 & 62.99 & 56.60 & 0.00 & 43.40 \\
    Ours (Qwen2.5-VL) & 45.6 & 64.62 & 20.10 & 33.30 & 46.60 \\
    Ours (Qwen2.5-VL) & 53.8 & 65.58 & 0.00 & 45.60 & 54.40 \\
    \bottomrule
  \end{tabular*}
\end{table}

The 53.8B average-active-parameter variant reaches 65.58\%, close to dense 72B baselines, while avoiding uniform reliance on the largest expert. Across routed settings, the controller assigns simple cases to lighter experts and reserves larger experts for more difficult cases. This supports the interpretation that the gain comes from task-adaptive expert selection, with orchestration overhead remaining below 12\% in the latency analysis.

\subsection{Expert-Usage Statistics by Benchmark}
\label{sec:appendix_expert_usage}

Table~\ref{tab:expert_usage_by_benchmark} reports how often the router selects the 32B and 72B Qwen2.5-VL experts across benchmarks. Easier general multimodal benchmarks are routed more often to the 32B expert, whereas more demanding video and holistic settings trigger a more balanced distribution. This provides a direct view of the computation allocation underlying the end-to-end scores.

\begin{table}[H]
  \centering
  \caption{Expert-selection frequency across benchmarks. Values report the percentage of routed visual expert calls assigned to each Qwen2.5-VL expert.}
  \label{tab:expert_usage_by_benchmark}
  \renewcommand{\arraystretch}{1.08}
  \scriptsize
  \setlength{\tabcolsep}{1.2mm}
  \begin{tabular}{lcc}
    \toprule
    \textbf{Benchmark} & \textbf{Qwen2.5-VL-32B (\%)} & \textbf{Qwen2.5-VL-72B (\%)} \\
    \midrule
    MME & 73.10 & 26.90 \\
    MMBench-EN & 69.80 & 30.20 \\
    MMStar & 71.80 & 28.20 \\
    LVBench & 52.40 & 47.60 \\
    MMMU & 63.60 & 36.40 \\
    Video-MME & 51.30 & 48.70 \\
    WorldSense & 54.20 & 46.80 \\
    \bottomrule
  \end{tabular}
\end{table}

\subsection{Direct Router Reliability Evaluation}
\label{sec:appendix_router_reliability}

We quantify whether the controller follows the closed-vocabulary routing schema by measuring valid control-token generation on held-out benchmark queries. We compare the training-free controller against a supervised router baseline obtained by LoRA fine-tuning Qwen2.5-14B on 7,500 successful multi-turn routing trajectories, with 1,200 held-out examples split by independent interaction sessions. The baseline uses LoRA rank 16, alpha 32, dropout 0.05, learning rate $2\times 10^{-4}$, global batch size 256, and a cosine scheduler; epoch 3 is selected from a 1--7 epoch sweep.

\begin{table}[H]
  \centering
  \caption{Closed-vocabulary routing success. Success means that the controller emits a valid, parseable control-token sequence in a single pass.}
  \label{tab:router_success}
  \renewcommand{\arraystretch}{1.08}
  \scriptsize
  \setlength{\tabcolsep}{0.8mm}
  \begin{tabular}{lccc}
    \toprule
    \textbf{Controller} & \textbf{Video-MME} & \textbf{WorldSense} & \textbf{Success (\%)} \\
    \midrule
    Supervised-trained & 2692/2700 & 3136/3172 & 99.20 \\
    Training-free (ours) & 2679/2700 & 3144/3172 & 99.10 \\
    \bottomrule
  \end{tabular}
\end{table}

The small 0.1\% gap indicates that the base controller's instruction-following ability is already sufficient for the routing protocol. Remaining invalid outputs are rejected by deterministic validation and fall back to a safe listen/text-only state, so invalid control strings do not execute expert calls.

\subsection{Interactive Multi-turn Evaluation}
\label{sec:appendix_multiturn_eval}

To better reflect real-time interaction, we restructure Video-MME questions into continuous three-round dialogue sessions per video. This setting tests shifting user goals, corrections, and memory reuse rather than isolated single-turn QA. Table~\ref{tab:interactive_multiturn} reports per-round accuracy, memory hit rate, and expert usage among cache misses.

\begin{table}[H]
  \centering
  \caption{Interactive three-round Video-MME evaluation. Expert usage is measured only among cache misses, because cache hits reuse committed memory records without invoking a visual expert.}
  \label{tab:interactive_multiturn}
  \renewcommand{\arraystretch}{1.08}
  \scriptsize
  \setlength{\tabcolsep}{0.7mm}
  \begin{tabular}{lcccc}
    \toprule
    \textbf{Round} & \textbf{Acc. (\%)} & \textbf{Memory Hit (\%)} & \textbf{32B Misses (\%)} & \textbf{72B Misses (\%)} \\
    \midrule
    Round 1 & 66.70 & 0.00 & 51.30 & 48.70 \\
    Round 2 & 69.20 & 96.20 & 71.20 & 28.80 \\
    Round 3 & 64.30 & 88.30 & 60.30 & 39.70 \\
    \bottomrule
  \end{tabular}
\end{table}

Memory reuse is strongest in the second round, where follow-up questions often refer to evidence already acquired in the first turn. The third round remains high but lower, reflecting more frequent topic shifts and corrections.

\subsection{Evidence-Key Robustness and Failure Cases}
\label{sec:appendix_failure_cases}

\noindent\textbf{Evidence-key design.}
The checksum component $h$ in Eq.~\ref{eq:keygen} is a strict SHA-1 checksum computed at the segment level, not a perceptual hash. Continuous media is partitioned before key generation, for example by VAD intervals for audio and by chunk/frame identifiers for video. Consequently, a local perturbation creates a local cache miss for the affected segment, while unchanged segments remain reusable. We choose exact matching to avoid erroneous reuse from fuzzy semantic matching; the trade-off is bounded local recomputation under minor input variation.

\noindent\textbf{Component instantiation.}
ASR requests use Whisper-Small. OCR and math requests are activated through routing tokens such as \texttt{[S.need\_ocr]} and \texttt{[S.need\_math]}; in our experiments these requests are served by the integrated multimodal expert pool consisting of Qwen2.5-VL-7B, Qwen2.5-VL-32B, and Qwen2.5-VL-72B.

\noindent\textbf{Failure modes.}
The main residual failure case is routing ambiguity when a query lacks a clear modality trigger. Such cases are uncommon: invalid or unresolved routing occurs in 0.8\% of Video-MME queries (21/2700) and 0.9\% of WorldSense queries (28/3172). The runtime handles them with a safe \texttt{[S.listen]} or text-only fallback, which may increase latency or produce a conservative no-op response but avoids invalid expert execution. Other practical limits include reduced responsiveness under frequent consecutive interruptions and degraded fine-grained temporal counting when event frequency exceeds the sparse frame-sampling rate.

\noindent\textbf{Latency and barge-in behavior.}
In the current unoptimized prototype, end-to-end throughput reaches 944.81 output tokens/s on a single NVIDIA H100 80GB GPU for queries with an average context length of about 512 tokens. The orchestration layer adds less than 12\% latency overhead, and retry rate remains below 1.2\%. During barge-in events, immediate cancellation prevents more than 90\% of scheduled compute from being spent on obsolete output or pending expert/TTS work.

\noindent\textbf{WorldSense Task Breakdown.} Table~\ref{tab:worldsense_breakdown} reports task-type breakdown on WorldSense, highlighting the performance gap between semantic/causal reasoning and fine-grained temporal tracking.

\begin{table}[H]
  \centering
  \caption{WorldSense task-specific breakdown showing performance variation across semantic reasoning vs. temporal tracking tasks. Overall accuracy is 44.07\% (Table~\ref{tab:omni_comparison}).}
  \label{tab:worldsense_breakdown}
  \renewcommand{\arraystretch}{1.10}
  \scriptsize
  \setlength{\tabcolsep}{1mm}
  \begin{tabular}{p{0.47\columnwidth} S[table-format=2.2] c}
    \toprule
    \textbf{Task Type} & \multicolumn{1}{c}{\textbf{Acc (\%)}} & \textbf{Correct/Total} \\
    \midrule
    \multicolumn{3}{l}{\textit{High-Performance (Semantic \& Causal)}} \\
    \addlinespace[0.15em]
    Emotion Change            & 60.42 & 58/96 \\
    Hallucination Detection   & 60.00 & 54/90 \\
    Temporal Prediction       & 58.18 & 64/110 \\
    Causal Reasoning          & 57.62 & 87/151 \\
    \addlinespace[0.3em]
    \multicolumn{3}{l}{\textit{Audio-Related (Omni-Modal)}} \\
    \addlinespace[0.15em]
    Audio Source Localization & 42.50 & 51/120 \\
    Audio Recognition         & 42.24 & 49/116 \\
    Audio Counting            & 34.44 & 31/90 \\
    \addlinespace[0.3em]
    \multicolumn{3}{l}{\textit{Challenging (Temporal Tracking)}} \\
    \addlinespace[0.15em]
    Spatial Relation          & 33.50 & 66/197 \\
    Temporal Localization     & 33.14 & 56/169 \\
    Object Counting           & 31.22 & 64/205 \\
    Action Counting           & 23.03 & 38/165 \\
    \bottomrule
\end{tabular}
\end{table}

\end{document}